%% file: full.tex
\documentclass{article}

\usepackage[utf8]{inputenc}
\usepackage[T1]{fontenc}
\usepackage{PRIMEarxiv}
\usepackage{authblk}
\usepackage{amsmath, amssymb, amsfonts, bm}
\usepackage{amsthm}
\usepackage{graphicx}
\usepackage{pdflscape}
\usepackage{placeins}
\usepackage{booktabs,longtable,array}
\usepackage{multirow}
\usepackage{caption}
\usepackage[dvipsnames]{xcolor}
\usepackage{tcolorbox}
\usepackage{url}
\usepackage{xurl}
\usepackage[numbers,square,sort&compress]{natbib}
\usepackage{fancyhdr}
\usepackage{microtype}
\usepackage{setspace}
\usepackage[left]{lineno}
\usepackage{xr-hyper}
\usepackage{hyperref}
\providecommand{\eprint}[2][]{%
  \href{https://arxiv.org/abs/#2}{\nolinkurl{#2}}%
}

\hypersetup{
  pdftitle={Energy-efficient operation of neural operators for virtual sensing},
  pdfauthor={Jason Yoo, Samrendra Roy, Souvik Chakraborty, Syed Bahauddin Alam},
  pdfkeywords={virtual sensing, neural operators, physical-field reconstruction, energy efficiency, near-sensor computing, edge deployment, nuclear energy systems},
  breaklinks=true,
  colorlinks=true,
  citecolor=Blue,
  linkcolor=Blue,
  urlcolor=Blue
}

\microtypecontext{spacing=nonfrench}

\graphicspath{{figures/}}

\input{sections/00_macros}

\title{Energy-efficient operation of neural operators for virtual sensing}

\author[1]{Jason Yoo}
\author[1]{Samrendra Roy}
\author[1,3,4]{Souvik Chakraborty}
\author[1,2,*]{Syed Bahauddin Alam}

\affil[1]{Nuclear, Plasma \& Radiological Engineering Department, Grainger College of Engineering, University of Illinois Urbana-Champaign, Urbana, IL, USA}
\affil[2]{National Center for Supercomputing Applications, Urbana, IL, USA}
\affil[3]{Department of Applied Mechanics, Indian Institute of Technology Delhi, New Delhi, India}
\affil[4]{Yardi School of Artificial Intelligence, Indian Institute of Technology Delhi, New Delhi, India}
\affil[*]{Corresponding author \href{mailto:alams@illinois.edu}{alams@illinois.edu}}

\begin{document}

\clearpage
\setcounter{page}{1}

\maketitle
\pagestyle{fancy}

\begin{abstract}
\noindent
\begin{tcolorbox}[
  colback=blue!10!white,
  colframe=blue!50!black,
  left=0mm,
  right=0mm
]
\noindent
\input{sections/abstract}
\end{tcolorbox}
\end{abstract}

\input{sections/01_introduction}
\input{sections/03_results}

\FloatBarrier

\input{sections/04_discussion}
\input{sections/02_methods}
\input{sections/06_backmatter}

\bibliographystyle{naturemag}
\bibliography{references}

\clearpage

\input{sections/07_extended_data}

\clearpage

\setcounter{table}{0}
\setcounter{figure}{0}
\setcounter{section}{0}
\setcounter{subsection}{0}

\renewcommand{\theHtable}{supp.table.\arabic{table}}
\renewcommand{\theHfigure}{supp.figure.\arabic{figure}}
\renewcommand{\theHequation}{supp.eq.\arabic{equation}}

\renewcommand{\theHsection}{supp.sec.\arabic{section}}

\renewcommand{\theHsubsection}{%
  supp.sec.\arabic{section}.subsec.\arabic{subsection}%
}

\renewcommand{\theHsubsubsection}{%
  supp.sec.\arabic{section}.subsec.\arabic{subsection}.subsubsec.\arabic{subsubsection}%
}

\input{si/05_si}

\end{document}

%% file: sections/00_macros.tex
\newcommand{\OpCert}{OpCert}

%% file: sections/abstract.tex
Virtual sensing repeatedly reconstructs physical fields from changing observations, often on a fixed geometry. We investigate how shared spatial computation reduces the energy of these updates while retaining the selected checkpoint and its evaluated predictions. In a heat-exchanger service, standard compiler freezing and explicit trunk reuse give similar operating energy reductions relative to graph replay: approximately 1\% at one request per second and 20\% at forty. In 15\,W mode with fixed clocks, reuse with graph replay completes the same request sequence with 22.0--22.5\% less energy than eager execution, including preparation and waiting. DeepONet and Fourier neural operator (FNO) controls distinguish the effects of reusable arithmetic and launch overhead. Preparation, artifact construction and worker replacement add costs outside repeated inference. These results connect operator structure to operating energy and show how update frequency and execution lifetime govern the benefit of computation reuse in physical-field virtual sensing.

%% file: sections/01_introduction.tex
Virtual sensing reconstructs physical quantities at locations that are difficult to instrument. Neural operators support this task by mapping accessible observations to spatial fields, including internal flow, heat transfer and component conditions in nuclear and other energy systems \citep{li2021fno,lu2021deeponet,hossain2025virtual,kobayashi2026mimonet,howes2026virso}. Proposed microreactor digital twins place predictive computation at both the reactor and the remote control room \citep{stevens2023digitaltwin}. The geometry and output grid often remain fixed while observations change. Each update therefore combines new input-dependent work with spatial computation shared across observations.

Once a predictor has been evaluated, its operating cost can be reduced while retaining the same predictions. Nuclear-industry guidance identifies power, latency and availability as operational considerations \citep{iaea2025aideployment}. Scientific verification and validation connect the model and its implementation to the intended use \citep{jakeman2026vv}. We hold the checkpoint, batch size of one and evaluated output fields fixed to isolate the effect of execution choices on energy. Reusing input-independent computation leaves each evaluated field update unchanged; reducing update frequency changes the observations represented by the available field.

Established methods address different sources of inference cost. Sparse activations and distillation change the predictor's computational structure \citep{howes2026sparse}, while deployed measurements show that algorithmic sparsity need not reduce energy \citep{yoo2026vswno}. For branch--trunk operators, partial evaluation and trunk precomputation retain the spatial representation determined by fixed query coordinates \citep{Jones1993PartialEvaluation,winovich2025active}. BLISSNet uses this approach for repeated flow reconstruction from sparse measurements \citep{veremchuk2026blissnet}. Standard compiler freezing similarly evaluates constant expressions before service \citep{pytorch2026freeze}. Model parameters remain fixed in every path; compiler freezing additionally transforms the execution graph. Graph replay reduces launch and allocation work while potentially retaining the full spatial computation \citep{pytorch2021cudagraphs,pytorch2026cudasemantics}.

These transformations exchange repeated work for retained storage and preparation. Their operating benefit depends on how often the representation is used and how long execution remains available. Our companion study compares neural-operator families, precisions and runtimes during continuous inference with preloaded inputs \citep{yoo2026energyvalidity}. Here we measure the energy of delivering the same complete field-update sequence, including preparation, data transport and waiting. We implement the comparisons in \OpCert{}, following established practices for output comparison, deployment testing and service recovery \citep{nvidia2026polygraphy,tensorflow2024infravalidator,ray2025serve}. Workers qualify against independently retained outputs before serving, and replacement workers use the same target. Execution paths share the service interface and numerical checks.

The heat-exchanger study connects reusable spatial computation to field predictions and inference energy (Fig.~\ref{fig:source_edge}), then measures complete service operation under clock and power policies (Fig.~\ref{fig:edge_sustained}). Darcy DeepONet and the Fourier neural operator (FNO) provide structural controls (Fig.~\ref{fig:service}). Matched comparisons with standard freezing resolve the effects of request rate, service duration and preparation cost (Fig.~\ref{fig:operating_cost}). Worker replacement and recorded fuel-cell measurements then relate rebuilding and update frequency to the fields available to the consumer. The Supplementary Information opens with a map of these comparisons and their measurement boundaries.

%% file: sections/03_results.tex
\section{Results}

\subsection{Reusing a fixed spatial representation reduces inference energy}

We use the released deterministic MIMONet heat-exchanger checkpoint to reconstruct pressure and three velocity components at 3,977 nodes from inlet conditions and a heat-flux profile \citep{kobayashi2026mimonet}. The checkpoint, float32 configuration, conversion to physical units and batch size of one remain fixed. The 310 test cases provide CFD reference fields for reconstruction assessment.

On Jetson, mean channel errors against CFD range from 0.52\% to 1.45\%, and the largest per-case relative $L_2$ error increase from the workstation evaluation is $9.16\times10^{-6}$ percentage points (Supplementary Table~\ref{tab:source-edge-errors}). All 310 outputs differ in bytes between platforms. Subsequent execution comparisons use the device-local eager reference.

For input $x_i$ and fixed query coordinates $X$, the branch--trunk computation has the form
\begin{equation}
 y_i=\mathcal C\!\left(B_\theta(x_i),T_\theta(X)\right),
 \label{eq:spatial_reuse}
\end{equation}
where $\theta$ denotes the retained weights, $B_\theta$ the input-dependent branches and $\mathcal C$ their contraction with the spatial representation $T_\theta(X)$. Partial evaluation moves this representation outside the repeated request path \citep{Jones1993PartialEvaluation,winovich2025active}. If its evaluation costs $C_T$ floating-point operations, replacing $N$ evaluations by one removes $(N-1)C_T$ operations. A representation with $n_q$ query points, feature width $p$, $c$ channels and $b$ bytes per element occupies $M_T=b n_qpc$ bytes. This counts the invariant tensor and arithmetic; preparation checks, capture and allocator storage contribute additional measured costs.

\begin{figure}[!htbp]
\centering
\includegraphics[width=\linewidth]{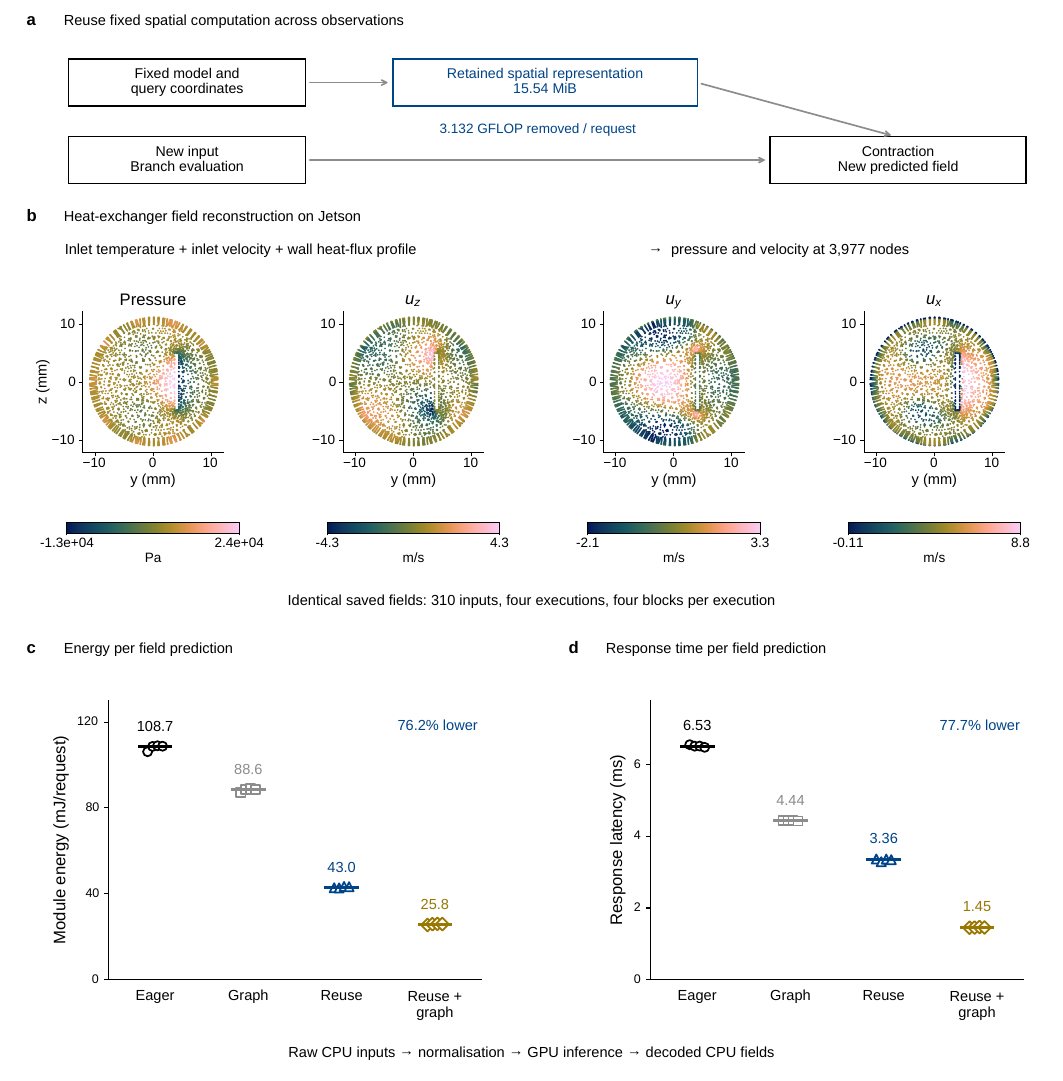}
\caption{\textbf{Trunk reuse reduces heat-exchanger inference energy.}
\textbf{a}, The spatial representation is retained; branches and contraction process each new input.
\textbf{b}, Predicted pressure and velocity at all 3,977 released $y$--$z$ query coordinates for CFD test case 1244, shown without spatial interpolation.
\textbf{c,d}, Module energy per request and block-median response latency for eager (E), graph (G), trunk reuse (R) and reuse with graph (RG). Points are fresh-process blocks ($n=4$ per path); horizontal marks give medians. Reductions are medians of paired ratios to E. Each path includes raw CPU input and decoded CPU output. Energy covers 20\,s of continuous inference at the retained MAXN configuration (Supplementary Note~\ref{si:source-edge}), without idle subtraction. All 310 saved fields agree with the device-local reference before and after decoding.}
\label{fig:source_edge}
\end{figure}

For the heat exchanger, retaining the $3977\times256\times4$ float32 representation requires 15.54\,MiB and removes approximately 3.132\,GFLOP of dense trunk computation per subsequent request. The branches and contraction still process each new input. We separate this reuse from graph replay using four paths on Jetson: eager, graph, reuse and reuse with graph (Fig.~\ref{fig:source_edge}).

Across four fresh-process blocks per path, all 16 saved output banks were byte-identical on the 310 cases before and after physical decoding. Reuse therefore retained the predicted fields and their existing errors against the deposited CFD targets.

Including raw CPU input and decoded CPU output, median module energy per request was 108.7, 88.6, 43.0 and 25.8\,mJ for eager, graph, reuse and reuse with graph, respectively. Paired reductions relative to eager were 18.4\% for graph, 60.2\% for reuse and 76.2\% for their combination; the combined response-time reduction was 77.7\%.

Lower energy per prediction did not require lower average power. Reuse with graph increased median mean module power from 16.58 to 17.51\,W but completed more requests in the 20\,s measurement interval. Peak reserved CUDA memory increased from 82 to 140\,MiB. These 20\,s continuous-inference measurements use the MAXN configuration detailed in Supplementary Note~\ref{si:source-edge}.

Standard compiler freezing can remove the same fixed-coordinate work. In an RTX~5060~Ti control, tracing and freezing a module with constant query coordinates reduced twelve Linear operations to the eight branch operations. Both freezing and manual reuse reproduced the eager bytes on all 310 inputs in three fresh processes (Methods; Supplementary Note~\ref{si:edge-service}).

\subsection{Energy savings persist over complete service operation}

We compare eager (E) and reuse with graph (RG) through the same worker transport, startup qualification and live guards. Both receive raw CPU inputs and return complete normalised and physically decoded arrays against the archived Jetson eager target. The offered rates span light load and operation near eager saturation.

\begin{figure}[!htbp]
\centering
\includegraphics[width=\linewidth]{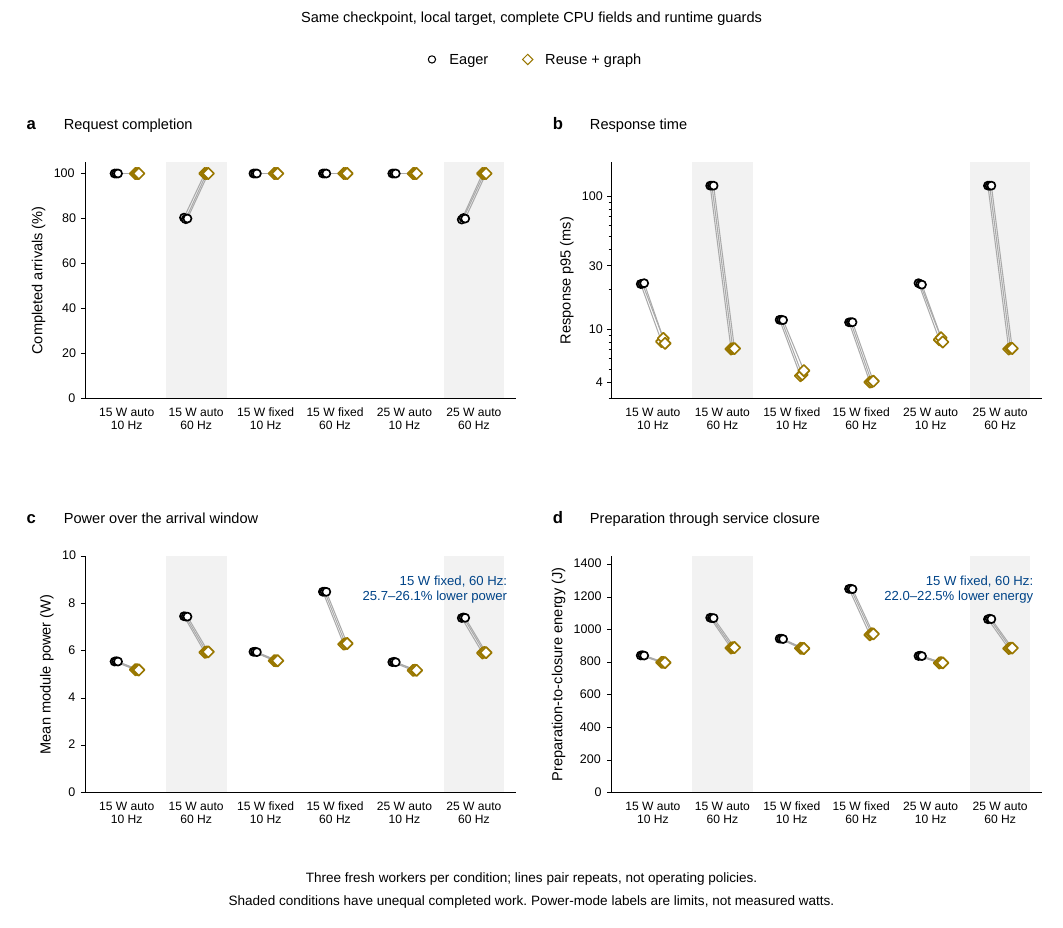}
\caption{\textbf{Energy savings persist over complete service operation.}
\textbf{a--d}, Completion fraction, response-time p95, mean module power during arrivals, and energy from preparation through closure. E and RG use the same transport and numerical checks. Points are fresh-worker blocks; lines pair repeats ($n=3$ per condition). Both paths complete all 7,200 arrivals at 15\,W fixed clocks and 60\,Hz. Shaded conditions have unequal completed work. Mode labels denote configured power policies. Total energy includes the 20\,s warmup and waiting, without idle subtraction; input-bank loading and cooling are excluded.}
\label{fig:edge_sustained}
\end{figure}

Thirty-six fresh-worker blocks cover 15\,W automatic clocks, 15\,W fixed clocks and 25\,W automatic clocks at 10 and 60 requests per second. All 142,551 returned requests matched the reference in post-delivery comparisons; 8,649 of the 151,200 arrivals exceeded the queue-age limit and were refused (Fig.~\ref{fig:edge_sustained}).

At 15\,W fixed clocks and 60\,Hz, both paths completed all 7,200 arrivals in every block. Median mean module power fell from 8.50 to 6.30\,W and response-time p95 from 11.27 to 4.01\,ms. Energy from preparation through closure decreased by 22.0--22.5\% across the three paired blocks, including warmup and waiting. Extended Data Fig.~\ref{ed:energy_boundary} resolves the energy by service phase.

The benefit depends on load and clock policy. At 10\,Hz, both paths completed every request and power differences were smaller. At 60\,Hz with automatic clocks, eager completed 5,739--5,790 requests in 15\,W mode and 5,727--5,777 in 25\,W mode, whereas RG completed all 7,200. These comparisons have unequal completed work. Fixing clocks allowed eager to complete the full workload; for RG, it reduced median p95 from 7.15 to 4.01\,ms but increased total energy by 9.08--9.61\%. Faster responses thus carried an energy cost.

\begin{samepage}
\subsection{The reusable work depends on operator structure}

Darcy DeepONet and FNO test whether the energy reduction follows the reusable computation rather than graph replay alone. Each comparison holds its checkpoint, batch-one inputs and device-local outputs fixed. Darcy DeepONet has the form of Eq.~\ref{eq:spatial_reuse}. Its branches process the changing input, while the trunk uses a fixed query grid. Retaining the trunk representation preserves the original query chunks and contraction order. Instrumented traces show dense matrix operations falling from 28 to four per request while six contractions remain: approximately 58.6\,GFLOP of repeated work is removed for 86.5\,MiB of storage (Extended Data Fig.~\ref{fig:physical}a).

\end{samepage}

The matched service cohort covers five DeepONet paths and three FNO paths on an embedded device and two workstation GPUs, with complete normalised CPU inputs and outputs. All 72 blocks completed; 21,888 replies matched the retained references, and 1,152 saved exemplars passed independent array comparison. A separate physical audit retained both models' existing prediction errors in all 80 decoded comparisons (Extended Data Fig.~\ref{fig:physical}b; Supplementary Notes~\ref{si:service} and~\ref{si:physical}).

\begin{figure}[!htbp]
\centering
\includegraphics[width=\linewidth]{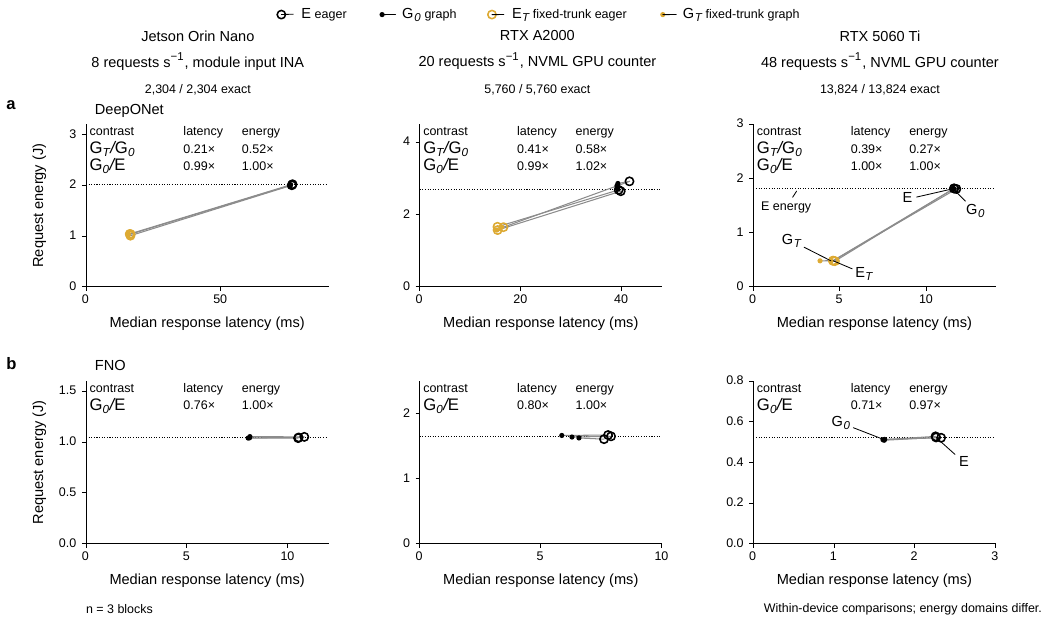}
\caption{\textbf{The energy benefit depends on operator structure.}
DeepONet (\textbf{a}) and FNO (\textbf{b}) services on Jetson, A2000 and RTX~5060~Ti, from left to right. Points show block-median response latency and interval energy per completed request ($n=3$ per arm and device); grey lines join paired blocks. Printed values are median paired ratios: $G_T/G_0$ for trunk reuse and $G_0/E$ for graph replay. Offered rates and energy domains differ between devices, so comparisons are within each device. All returned outputs match the local reference. Header counts include $G_H$, omitted from the plot; complete costs and graph-baseline contrasts are in Supplementary Tables~\ref{tab:si-control-costs} and~\ref{tab:si-control-contrasts}.}
\label{fig:service}

\end{figure}

Adding trunk reuse to an already prepared graph service reduced paired request energy by 48.36\% on Jetson, 41.84\% on A2000 and 73.49\% on RTX~5060~Ti (Fig.~\ref{fig:service}a). These comparisons share the supervisor, transport and numerical checks. Fixed-trunk graph reduced response time by 61--80\% relative to eager, but adding graph replay after reuse did not consistently improve energy (Supplementary Table~\ref{tab:si-control-contrasts}).

FNO has no corresponding fixed trunk. Native graph replay reduced paired response time by 20--29\%, whereas median paired request-energy changes were below 3\%, with paired ranges crossing parity on A2000 and Jetson (Fig.~\ref{fig:service}b). The additional constructor-hoisting control retained more memory without consistently improving response time beyond native capture (Supplementary Note~\ref{si:qualification}). Energy includes interarrival waiting, so the latency reduction need not produce a proportional saving. Comparisons are within each device because offered rates, telemetry domains and application co-use differ (Supplementary Note~\ref{si:service}).

\subsection{Standard freezing and the cost of retaining execution}

The arithmetic saved per request must be considered alongside preparation and waiting. For execution path $a$, request rate $\lambda$ and arrival horizon $T$, we partition preparation-to-closure energy as
\begin{equation}
 E_a(\lambda,T)=S_a(\lambda,T)+T\,\overline P_a(\lambda,T),
 \label{eq:service_energy}
\end{equation}
where $\overline P_a$ is mean measured power over the full arrival horizon, including waiting, and $S_a$ is the energy in the remaining phases: preparation, any prescribed ready-idle or warmup interval, drain and closure. For $N=\lambda T$ arrivals all completed, energy per prediction is $E_a/N=S_a/N+\overline P_a/\lambda$. Both terms are measured for each condition. This accounting separates costs outside the arrival window from the energy of supplying the request sequence; one-time frozen-artifact construction is reported separately.

\begin{figure}[p]
\centering
\includegraphics[width=\linewidth]{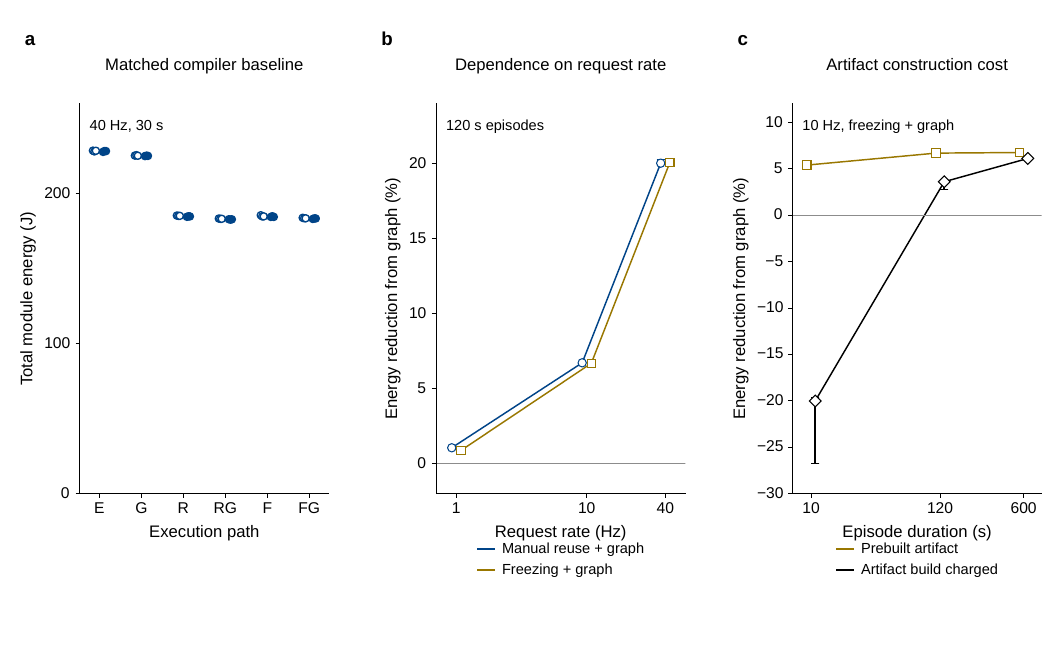}
\caption{\textbf{Standard freezing obtains the reuse benefit, subject to operating costs.}
\textbf{a}, Preparation-to-closure energy for six heat-exchanger paths at 40\,Hz over 30\,s. Open and filled points distinguish two sessions, each with three fresh-worker blocks per path; bars mark medians. F denotes standard freezing; other paths follow Fig.~\ref{fig:source_edge}.
\textbf{b}, Paired energy reductions relative to native graph over 120\,s at each request rate.
\textbf{c}, Freezing with graph at 10\,Hz, with the artifact prebuilt or construction charged to each episode. Points in \textbf{b,c} are medians of three episode pairs. The construction charge uses the median of five fresh-process builds; whiskers span the observed episode pairs and, where charged, all five build costs. Lines guide the eye; negative reductions denote increased energy. All service blocks use 15\,W automatic clocks, complete equal work and include preparation, waiting and closure without idle subtraction.}
\label{fig:operating_cost}
\end{figure}

We next compare standard freezing (F) and manual trunk reuse (R), each with and without graph replay, in the same heat-exchanger service. All paths use 15\,W automatic clocks, with qualification and capture during preparation and no additional warmup. At 40\,Hz over 30\,s, all six paths complete 1,200 requests per block. Across two sessions, manual reuse with graph reduces total energy by 18.7--18.8\% relative to native graph; freezing with graph gives 18.5--18.6\% (Fig.~\ref{fig:operating_cost}a). Standard freezing thus obtains a similar benefit without the manual trunk rewrite. Each session contains three paired blocks per comparison.

Request frequency changes the realised saving. Over 120\,s, manual reuse with graph saves 1.06\% at 1\,Hz, 6.73\% at 10\,Hz and 20.05\% at 40\,Hz relative to native graph. Freezing with graph gives 0.88\%, 6.70\% and 20.09\%, respectively (Fig.~\ref{fig:operating_cost}b). All paths complete the same work within each condition. Independent ready-idle measurements give approximately 4.61\,W for all three paths; this common module cost limits the benefit at light load.

Artifact construction changes the comparison for short deployments. Five fresh-process builds require 19.62--24.96\,J and 3.42--4.45\,s, with medians of 19.79\,J and 3.46\,s. Every saved artifact reproduces all 310 reference fields before and after decoding. At 10\,Hz, using a prebuilt artifact saves 5.41\%, 6.70\% and 6.77\% over 10, 120 and 600\,s. Charging the median build cost to each episode changes these values to a 20.01\% increase, a 3.62\% saving and a 6.13\% saving (Fig.~\ref{fig:operating_cost}c). Every measured build costs more than the 4.10--4.28\,J saved over 10\,s and less than the 42.50--43.04\,J saved over 120\,s. The sign of the comparison therefore holds across the observed build and episode ranges. Subsequent workers can use the same artifact without rebuilding it. Supplementary Note~\ref{si:edge-service} separates artifact construction from worker preparation; Tables~\ref{tab:matched-freezing} and~\ref{tab:operating-lifetime} report complete episode costs.

The Darcy and FNO controls also include preparation and closure. At 1\,Hz, adding trunk reuse to the Darcy graph service reduces total energy by 16.98\%. FNO graph replay increases total energy by 0.14\% at 1\,Hz and reduces it by 0.49\% at 8\,Hz relative to eager. The incomplete Darcy 8\,Hz comparison is excluded from energy contrasts; its queue refusal and subsequent preparation failure are retained in Supplementary Note~\ref{si:edge-service}.

\FloatBarrier

\clearpage
\subsection{Retaining computation differs from holding an earlier field}

Computation reuse supplies a fresh prediction for each processed observation. Holding an earlier output instead changes the field available at that time. We examine this distinction using 2,642 recorded fuel-cell observations and two fixed predictors with the same inputs. Refreshing once per 60 observations increases temperature RMSE from 0.03756 to 0.11279\,$^\circ$C for MIMONet and from 0.01120 to 0.11226\,$^\circ$C for ridge. The difference between the fresh predictors nearly disappears under this policy (Supplementary Fig.~\ref{sifig:temporal_scope}d,e). Errors use 292 grid locations excluded from the inputs; temperature fields inherit the released interpolation of 81 measured channels. These offline policies measure field error; their energy was not measured. A separate fuel-cell replay evaluates held fields during worker replacement (Supplementary Note~\ref{si:temporal-scope}).

Worker replacement also interrupts field delivery. Twelve heat-exchanger traces apply power-mode changes and either a tracked weight mutation or worker exit. All 24 transitions between 15 and 25\,W retain the worker and its numerical outputs. After the faults, all twelve replacement workers qualified against the archived target in 3.78--4.63\,s. Of 7,200 arrivals, 6,631 returned matching outputs, 503 were unavailable during recovery and 66 exceeded the queue-age limit (Extended Data Fig.~\ref{fig:edge_recovery}).

Ray Serve, supplied with the same reused graph backend, CPU interface, witnesses and guards, also recovered the original target. Its configured transport and controller produced different operating costs and update availability (Supplementary Fig.~\ref{sifig:service_comparison}). Controlled allocator histories and recovery ablations resolve the roles of execution state and replacement in Supplementary Notes~\ref{si:history} and~\ref{si:contractfrontier}.

%% file: sections/04_discussion.tex
\section{Discussion}

Operator structure determines the computation that can be shared across observations; service policy determines the resulting energy saving. In the heat-exchanger study, standard freezing and explicit trunk reuse remove the same fixed-coordinate work and achieve similar operating costs. Their measured reduction relative to native graph replay grows from approximately 1\% at 1\,Hz to 20\% at 40\,Hz. A retained spatial representation therefore has different value under occasional updates and sustained use, even for the same checkpoint and output fields.

The structural controls explain this dependence. DeepONet removes repeated trunk arithmetic while processing each new input through its branches and contraction. Its retained storage grows with query count and feature width. FNO graph replay reduces launch overhead and response latency, with little change in paced-service energy. The paced comparisons include interarrival waiting; the complete-episode controls also include preparation. Clock policy introduces another trade-off: fixed clocks shorten heat-exchanger responses but increase RG service energy by approximately 9\%. Execution selection consequently requires the cost of completing the intended update sequence, together with its response times and memory requirements.

Preparation matters when execution is short-lived or rebuilt. The measured frozen-artifact construction cost exceeds the saving from the 10\,s episode and falls below the savings from the 120 and 600\,s episodes. Construction is charged once per artifact; preparation is incurred at each worker start. An artifact can serve multiple worker lifetimes while its checkpoint, query geometry and execution configuration remain fixed. 

Worker replacement also interrupts delivery. Both OpCert and Ray Serve recover the retained numerical target; their configured transport and recovery procedures produce different costs and unavailable updates. The fuel-cell record shows why update availability requires its own assessment. The advantage of the more accurate fresh temperature predictor nearly disappears when both predictors hold their outputs for 60 observations. Reusing input-independent computation continues to evaluate new observations, whereas reducing field updates changes the available physical information.

The comparisons use deterministic, fixed-query models and device-local outputs evaluated on finite input collections. Preserving those arrays retains their existing physical-field errors. The source-to-edge comparison assesses the separate numerical change introduced by the platform transition. Heat-exchanger reconstruction is evaluated against CFD, and the recorded fuel-cell sequence provides a timing control with a more accurate ridge baseline. Application assessment additionally includes sensors, uncertainty and update requirements \citep{jakeman2026vv,iaea2025aideployment}. For a specified field-update sequence, the results support selecting execution by reusable work, response requirements and the lifetime over which preparation is amortised. Established transformations can then reduce energy while retaining the selected predictions and processing each new observation.

%% file: sections/02_methods.tex
\section*{Methods}
\label{sec:methods}

\subsection*{Heat-exchanger model and execution controls}

We use the released deterministic heat-exchanger checkpoint and its native 3,977-node data without retraining \citep{kobayashi2026mimonet,kobayashi2026mimonetdata}. The reference uses deterministic single-pass inference. The published uncertainty model uses 20 dropout-sampled evaluations of both branches and trunk; uncertainty calibration is outside this comparison. Outputs are pressure, $u_z$, $u_y$ and $u_x$; inputs are an inlet-condition pair and a 100-point heat-flux profile. The released evaluator fits normalisation from the first 1,082 of 1,546 cases, skips 154 and tests the final 310. Coordinate scaling, input normalisation and the CPU inverse scaler are unchanged. We reproduce batch-eight evaluation before retaining a batch-one workstation reference. Supplementary Note~\ref{si:source-edge} records the software versions, field errors and resource costs.

The heat-exchanger experiments use float32 inference, one PyTorch CPU thread, deterministic algorithms and deterministic cuDNN selection, no cuDNN benchmarking or TF32, and \texttt{CUBLAS\_WORKSPACE\_CONFIG=:4096:8}. Fixed-trunk reuse retains the original $1\times3977\times256\times4$ float32 trunk tensor; both branches and the original contraction remain dynamic. The separate single-process factorial compares eager, graph, reuse and reuse with graph in four cyclic temporal groups. Each fresh-process block has 10\,s warmup and a 20\,s continuous measurement window. One outstanding request includes raw CPU input, normalisation, GPU transfers/execution and complete decoded CPU output. Startup and post-block byte/CFD comparisons lie outside that interval. The 108,266 timed requests cycle through the same 310 inputs. Complete normalised and decoded banks are saved after every block.

The isolated factorial reports execution costs even when the workstation byte criterion is not met. Their local output agreement is reported separately from source admission. Reapplying both CPU decoder stacks to the same normalised source output diagnoses a separate decoding boundary. The factorial retained the existing MAXN\_SUPER configuration because the requested mode change required reboot. Its costs therefore apply to that recorded configuration. Exact clock bounds, TPC masks and source/edge versions remain in Supplementary Note~\ref{si:source-edge}. The later service below uses the recorded configuration in which the tested mode changes succeed without reboot.

\subsection*{Matched heat-exchanger service and operating changes}

Before measuring service, we fix the checkpoint, raw input bank, decoder, execution code and Jetson eager reference. Eager (E) and fixed-trunk graph (RG) share one backend and worker transport. Each request carries the full raw float64 CPU input, applies frozen float64 normalisation before float32 inference, and returns complete normalised and physically decoded float32 CPU fields. Positions 0--7 supply twice-repeated startup witnesses; positions 8--309 are cycled in warmup and service. These inputs are exercised before timing. Returned fields are compared with their archived answers after delivery, with comparison and logging inside the instrumented workload.

Three fresh-worker blocks per arm cover 15\,W automatic clocks, 15\,W fixed clocks and 25\,W automatic clocks, each at 10 and 60\,Hz. A block has 20\,s warmup and a 120\,s arrival horizon. Arrivals over 100\,ms old at dispatch are refused and retained in accounting. The offered rates span light load and eager saturation; the queue limit is an experimental setting rather than a plant deadline. Cooldown lasts at least 20\,s and requires available temperature channels at or below $45\,^{\circ}$C for 5\,s, with a 180\,s limit; unavailable channels remain recorded.

Execution comparisons pair replicate identifiers within a mode. The 15\,W clock-policy groups alternate their order; 25\,W is a separate campaign. Fixed clocks affect CPU, GPU and memory together, and the modes also differ in CPU limits. Thus cross-mode records do not isolate a GPU-clock or power-cap effect. Equal-work energy contrasts require completion of the same entire offered sequence. Timestamped module-power integration separates preparation, warmup, arrivals, draining and closure. Preparation-to-closure energy includes warmup and instrumentation, but excludes input-bank loading and inter-block cooling. The offline audit verifies recorded comparisons, executed code and request accounting; full arrays were not retained for every reply.

Twelve additional 60\,s traces at 10\,Hz test E and RG with either tracked weight mutation or worker exit, three traces each. Modes change from 15 to 25\,W at 10\,s and back at 20\,s without replacing the worker; the fault occurs between requests at 30\,s. The controller then quarantines and explicitly replaces it. Process identities, generations and reference hashes distinguish handoffs from independent launches. Recovery energy is nested within trace energy, not added twice. Unavailable arrivals during these injected faults are retained in the accounting in Supplementary Note~\ref{si:edge-service}.

\subsection*{Standard freezing and operating costs}
The matched extension uses 15\,W automatic clocks, the archived device-local target and the same CPU interface, startup witnesses and live guards. Six heat-exchanger paths (E, G, R, RG, F and FG) run at 40\,Hz for 30\,s in two separated sessions, each with three fresh-worker blocks per path. Cyclic arm orders cover all six starting positions across the sessions. G, RG and FG additionally run at 1, 10 and 40\,Hz for 120\,s, and at 10\,Hz for 10 and 600\,s, with three cyclically ordered repeats per condition. These are complete episodes, not prefixes of longer runs. Nine controls measure 60\,s of ready-idle power. No discretionary warmup is added; qualification and graph capture remain part of preparation. Between blocks, cooling lasts at least 20\,s and requires monitored temperatures below $45\,^{\circ}\mathrm{C}$ for 5\,s, with a 180\,s timeout.

Module-input power is sampled at a nominal 20\,ms interval and integrated over preparation through worker closure, including the full arrival horizon and drain. Input-bank loading and cooling are excluded. Five additional fresh-process builds measure artifact construction with the same builder, checkpoint and 15\,W automatic-clock configuration. The interval includes library imports, input and model loading, CUDA initialisation, tracing/checking, freezing and saving, but excludes interpreter startup and post-build evaluation. A separate process samples module power; the maximum observed gap is 20.6\,ms. Each saved artifact is reloaded and compared with all 310 archived normalised and decoded outputs outside the energy interval. Eighteen preflight banks cover all 310 inputs in both normalised and physical representations; full byte comparison is independent of the runtime post-delivery audit. The extension acquired 115 of 126 planned blocks: 90 heat-exchanger blocks, 12 complete Darcy 1\,Hz blocks, one incomplete Darcy 8\,Hz block and 12 complete FNO blocks. Unequal completed work is excluded from paired energy estimates. Supplementary Note~\ref{si:edge-service} reports the failed continuation, measurement limits and structural controls.

\subsection*{Energy, timing and statistical units}
\label{methods:measurement}
\label{methods:service}

Embedded-device energy integrates timestamped module-input VDD\_IN samples by the trapezoidal rule without idle subtraction. The requested 20\,ms polling interval is not the underlying sensor resolution. Continuous inference, native-cadence replay and preparation-to-closure measurements use the separate intervals specified for each cohort. Startup gaps in the Ray heat-exchanger records approach one second, compared with below 28\,ms during service, so startup energy has lower temporal resolution.

For the Darcy paced-service cohort, request energy is
\[
 e_{\mathrm{request}}=N_{\mathrm{completed}}^{-1}
 \int_{t_0}^{t_1}P(t)\,\mathrm{d}t,
\]
where $N_{\mathrm{completed}}$ is the completed-request count and $P(t)$ is the measured power. The interval $[t_0,t_1]$ spans input encoding through completed delivery and at least the full arrival horizon, including interarrival idle. Discrete-GPU values use cumulative NVML energy-counter differences; Jetson values integrate module-input samples. These hardware domains differ \citep{nvidia2026nvml,nvidia2026jetsonpower} and do not provide a controlled cross-device efficiency ranking. Preparation, reference generation, auditing and recovery are outside this cohort's request-energy boundary and are reported separately. Observed application co-use is included without per-process attribution.

The experimental unit is a fresh process or worker block. Matched ratios are computed within recorded groups, then summarised by medians and observed ranges; these are not confidence intervals. Requests, telemetry samples and grid points are not independent replicates. Physical-field comparisons use matched timestamps and locations, and repeated replay runs of the same sequence do not create independent physical experiments.

\subsection*{Retained targets, qualification and runtime checks}
\label{methods:state}
\label{methods:qualification-record}
\label{methods:qualification}
\label{methods:preservation}
\label{methods:guards}

A checkpoint $\theta$ and numerical configuration $\sigma$ do not specify all worker-local state $s$. Execution maps input $x$ to output $y$ and next state $s'$, written $(y,s')=F_\theta(x;\sigma,s)$; $s$ includes owned storage, persistent tensors and runtime choices. We retain the numerical output of a designated execution outside the worker and compare replacements with that same target. The heat-exchanger service retains an archived Jetson eager bank; the earlier cross-platform comparison retains a separate workstation bank. Darcy references are device/configuration local. Supplementary Note~\ref{si:scope} records their checkpoint, input and environment provenance.

The qualification record contains the checkpoint, input/output interface, numerical configuration, reference corpus, comparison predicate, witnessed evidence, monitored conditions and recovery policy. Its evidence-linked schema and exported examples are described in Supplementary Note~\ref{si:contractfrontier}. The owner of an application must justify its reference and permitted differences. We use byte identity to hold the selected prediction fixed while changing its execution, and evaluate illustrative tolerance predicates in a separate experiment. Retaining the original reference also avoids changing a tolerance policy across replacements: $|c_1-r|\leq\varepsilon$ and $|c_2-c_1|\leq\varepsilon$ do not imply $|c_2-r|\leq\varepsilon$.

For retained qualification pairs $\{(w_j,r_j)\}_{j=1}^{m}$, each candidate worker $g$ evaluates every witness twice. With $C_g^{(k)}$ denoting repeat $k$, admission requires
\begin{equation}
 \mathcal E(C_g^{(k)}(w_j),r_j)=\mathrm{true},
 \qquad j=1,\ldots,m,\quad k=1,2.
 \label{eq:reference_contract}
\end{equation}
The heat-exchanger, fuel-cell and paced Darcy services use eight witnesses. Admission requires finite outputs and byte agreement: identical structure, shape, dtype and logical element bytes, including signed-zero representations. These comparisons establish numerical agreement on the witnessed executions. A tolerance predicate need not be an equivalence relation, and the selection among qualified candidates additionally depends on latency, energy, memory and service policy.

The worker verifies staged artifacts, prepares the execution and applies Eq.~\ref{eq:reference_contract} before admission. For the heat-exchanger and fuel-cell services, live guards cover exposed numerical settings, model/query/trunk identities and available tensor versions, normalisation/decoder state, schema and finiteness. The Darcy guards and allocator conditions are specified below. They do not inspect private library state or detect all untracked writes. Inference tensors can lack version counters, so retained storage also requires exclusive ownership \citep{pytorch2026inferencemode}. Each protocol identifies whether numerical comparison precedes or follows delivery.

Sufficient conditions for preservation over the declared input domain are the same deterministic arithmetic and ordering, invariant values from immutable dependencies, preserved data dependencies and storage ownership, and a fixed numerical configuration. Supplementary Note~\ref{si:qualification} gives the dependency-graph argument and empirical-pass limitations \citep{Jones1993PartialEvaluation,pnueli1998translation,necula2000translation}. The implementation does not mechanically verify these conditions for private runtime state; agreement on finite witnesses therefore remains distinct from preservation over the full input domain.

An error or timeout quarantines the worker. Recovery retires its process, checks the prescribed resource conditions and qualifies a replacement against the original target. Failed requests are not replayed automatically. Measured replacements use byte identity; relaxed-predicate experiments test startup admission, not recovery. Independently owned CPU responses prevent later GPU-buffer reuse from changing an already delivered field.

\subsection*{Darcy execution and physical-error controls}
\label{methods:workloads}
\label{methods:execution}
\label{methods:physical}

Retained Darcy DeepONet and FNO checkpoints \citep{lu2021deeponet,li2021fno} have native grids of $421\times421$ and $85\times85$. Each 24-input bank separates eight qualification witnesses from sixteen serving positions. Numerical configuration is float32 evaluation/inference mode with one CPU thread, cuDNN benchmarking and deterministic selection disabled, global deterministic enforcement disabled, cuDNN TF32 enabled, matrix-product TF32 disabled and \texttt{CUBLAS\_WORKSPACE\_CONFIG=:4096:8}. These settings differ from the heat-exchanger cohort and remain bound to the corresponding references.

The 72-block experiment compares five DeepONet paths (eager, prepared graph, fixed-trunk eager, fixed-trunk graph and empirical-pass graph) and three FNO paths (eager, native graph and empirical-pass graph). Preparing only the fixed $2\pi$ tensor permits the DeepONet graph control; fixed-trunk paths preserve query grids, chunks and contraction order. Native FNO graph capture requires no empirical substitutions. CUDA graphs and partial evaluation are established mechanisms \citep{pytorch2021cudagraphs,Jones1993PartialEvaluation}; empirical constructor hoisting is an attribution control, with its invariance limits documented in Supplementary Note~\ref{si:qualification}.

Complete normalised CPU tensors cross the worker boundary and independently owned CPU outputs return after synchronisation. Latency starts before input encoding and ends at complete reply receipt; response time additionally includes scheduled-arrival delay. Input acquisition, scientific preprocessing and physical decoding lie outside this boundary. Live checks occur before execution and after output materialisation. Observed allocation retries or OOM prevent preparation/admission/service; these counters indicate allocator events, not plan changes \citep{pytorch2026memorystats}. Serving inputs are checked twice before timing and are not unseen runtime tests. Three fixed execution orders per model/device have the balance limitations listed in Supplementary Note~\ref{si:service}. Fresh workers warm for at least 5\,s and a full input cycle; 12\,s arrival horizons offer 576, 96 and 240 requests on RTX~5060~Ti, Jetson and A2000, respectively.

Outputs remain in client memory until the energy interval ends, then all are compared with retained answers. One complete exemplar per serving input and block supports independent reloading; repeated requests otherwise retain equality flags and digests. Physical auditing separately decodes 80 matched candidate/reference pairs with the unchanged CPU inverse normalisation. For decoded prediction $d(y_j)$ and simulated target $u_j^*$, error is $\|d(y_j)-u_j^*\|_2/\|u_j^*\|_2$, evaluated in float64 with no denominator floor. Decoding and scoring lie outside the paced-service interval. Native grids are not regridded to rank architectures. Supplementary Note~\ref{si:physical} identifies audited arms and the separate Well persistence diagnostic.

\subsection*{Comparison with an established serving framework}

Ray Serve~2.49.2 runs on the same embedded device in an isolated Python environment retaining the installed PyTorch and NumPy versions \citep{ray2025serve}. Ray and OpCert share the RG backend, full CPU interface, numerical target, witnesses and guards. The shared application adapter supplies numerical qualification to both services. Each replica constructor qualifies against the unchanged target; Ray health checks mark a guarded replica unhealthy and its controller replaces it. OpCert explicitly invokes replacement after quarantine.

Three paired fresh-driver blocks offer 7,200 requests over 120\,s at 15\,W with automatic clocks, following 20\,s warmup. Both services use one replica, one outstanding request, one PyTorch thread and the same 100\,ms dispatch-age limit. Ray uses deployment handles without HTTP. Returned digests are compared after delivery, and each driver independently saves the complete 310-input output bank. A tracked mutation follows the service interval; recovery requires a new worker, qualification against the original target and an exact reply. Supplementary Note~\ref{si:edge-service} specifies resource allocations, health-check and retirement settings, acquisition order and the prospectively repeated pair.

\subsection*{Numerical alternatives and their scope}

The primary service comparisons use a fixed checkpoint, physical decoder and device-local byte target. Separate Darcy experiments evaluate elementwise, field, gradient and section-flux predicates on eight qualification and sixteen evaluation inputs, with budgets fixed before candidate comparison. Supplementary Note~\ref{si:contractfrontier} defines these predicates, their conditional error bound, candidate availability and live admission results. The compile/TensorRT cost comparisons use resident input-bank indices and different publication checks, and remain separate from the full-input services. These experiments do not establish application acceptance thresholds or recovery under relaxed predicates.

\subsection*{Resource-history and profiling controls}
\label{methods:history}
\label{methods:profiling}

The allocator intervention changes the process-visible PyTorch ceiling, rather than total free device memory. Caps use the workload-specific allocation and workspace scales recorded in Supplementary Note~\ref{si:history}; native-convolution controls follow the same operation across caps. Releasing unused allocator blocks need not clear the thread-local cuDNN plan cache \citep{pytorch2026convv8}. Cache-disabled and fresh-process controls therefore test distinct operations, always against the pre-intervention reference. External occupancy uses an independent CUDA helper before worker initialisation and is a separate intervention, not a proxy for the allocator cap.

Ownership controls compare eager, graph and a retained allocator pool under the same cap/reclaim sequence \citep{pytorch2026cudasemantics}. Allocation counters and request outcomes distinguish observed resource refusals from numerical differences. An $A$--$B$--$A$--$C$--$A$ input-history control additionally tests request association and preservation of earlier CPU replies under buffer reuse. Supplementary Notes~\ref{si:history} and~\ref{si:legacy} retain cap matrices, input modifications, failed trials and hardware-specific stopping conditions. The controls establish outcomes under the specified histories; they do not estimate their frequency in ordinary deployment.

Nsight Systems traces are collected separately from primary performance measurements \citep{nvidia2026nsys}. Kernel durations and graph-execution intervals are overlapping categories and are not added into an exclusive latency decomposition. The prepared-input seven-GPU study, 45-block service study and wider admission cohort retain their original interfaces and numerical configurations in Supplementary Notes~\ref{si:profiling} and~\ref{si:legacy}; their measurements are not pooled with the primary services.

\subsection*{Recorded-cadence and update-policy comparisons}

The fuel-cell replay and offline update-policy analysis use a chronological split, a ridge comparator, recorded timestamps and complete physical-field outputs. The released preprocessing takes the magnitude of $18\times18$ segment currents in A/segment and interpolates $9\times9$ measured temperatures to an $18\times18$ grid. Each predictor receives 32 selected grid locations per channel; the temperature inputs are derived observations rather than 32 independent thermometers. Supplementary Note~\ref{si:temporal-scope} specifies fitting, input populations, delivery and hold-last conventions, the prediction/execution/retention error decomposition, and the 120.998\,s common energy window. The offline policies assess field error without power measurements. The device replay measures interruption and energy over its own observation window.

%% file: sections/06_backmatter.tex
\section*{Data availability}
The data supporting this study, including figure data, retained output arrays, execution records and profiling traces, are available to editors and reviewers from the corresponding author on request during peer review. Public deposition is planned upon acceptance. The original MIMONet checkpoint and heat-exchanger simulation data are available from the archived data and code release \citep{kobayashi2026mimonetdata}; the measured fuel-cell data are deposited by the University of Seville \citep{toharias2024fuelcelldata}. The Well datasets and released checkpoints are available through the original project \citep{ohana2024well}.

\section*{Code availability}
The \OpCert{} implementation, experiment code and analysis scripts are available to editors and reviewers from the corresponding author on request during peer review. Reproduction instructions and the code versions used for the reported experiments are retained with these materials. Public code release is planned upon acceptance.

\section*{Acknowledgements}
The authors thank the National Center for Supercomputing Applications (NCSA) at the University of Illinois Urbana-Champaign for computational support through the Delta and DeltaAI systems.

\section*{Funding statement}
This research was supported by the National Science Foundation (NSF) under Award No. ECCS-2543177.

\section*{Author contributions}
J.Y. conceived and led the study, developed \OpCert{}, performed the experiments and analyses, and wrote the manuscript. S.R. contributed model resources and technical input. S.C. contributed model resources, provided research guidance, and reviewed the manuscript. S.B.A. supervised the work, acquired funding, and reviewed the manuscript.

\section*{Competing interests}
The authors declare no competing interests.

\section*{Additional information}
\textbf{Supplementary information} is supplied with this manuscript.
\textbf{Correspondence and requests for materials} should be addressed to S.B.A.
(\href{mailto:alams@illinois.edu}{alams@illinois.edu}).

%% file: sections/07_extended_data.tex
\clearpage
\begingroup
\makeatletter\setlength{\@fptop}{0pt}\setlength{\@fpbot}{0pt plus 1fil}\makeatother
\renewcommand{\figurename}{Extended Data Fig.}
\captionsetup[figure]{name=Extended Data Fig.}
\renewcommand{\thefigure}{\arabic{figure}}
\setcounter{figure}{0}
\ifdefined\theHfigure
  \renewcommand{\theHfigure}{extendeddata.\arabic{figure}}
\fi

\begin{figure}[p]
\centering
\includegraphics[width=\linewidth]{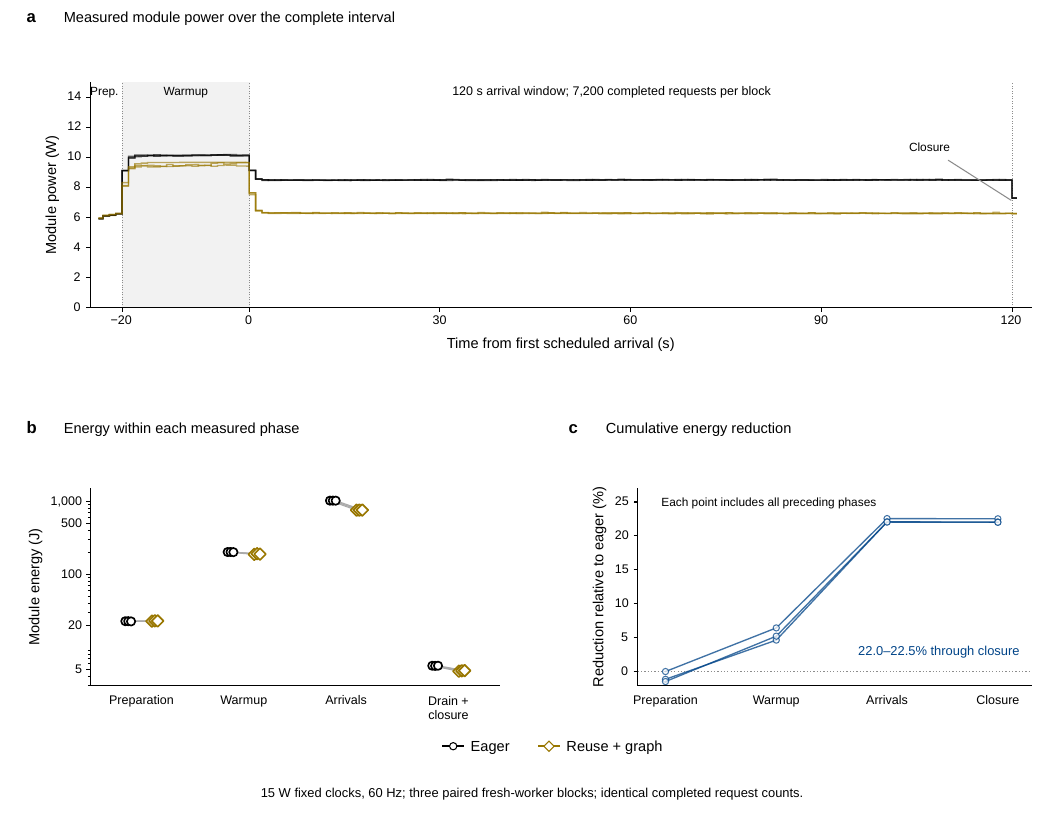}
\caption{\textbf{Energy saving includes preparation and warmup.}
\textbf{a}, Module power for three paired E and RG blocks at 15\,W fixed clocks and 60\,Hz. Curves are one-second time-weighted means, split at phase boundaries; time zero is the first arrival.
\textbf{b}, Energy by phase, integrated from the original samples, on a logarithmic axis. Points are blocks; lines pair repeats.
\textbf{c}, Cumulative paired reduction through each phase. All blocks complete 7,200 requests, with a final reduction of 22.0--22.5\%. Preparation includes qualification; the 20\,s warmup is retained. Energy is not idle-subtracted; bank loading and cooling are excluded.}
\label{ed:energy_boundary}
\end{figure}
\clearpage
\begin{figure}[t]
\centering
\includegraphics[width=\linewidth]{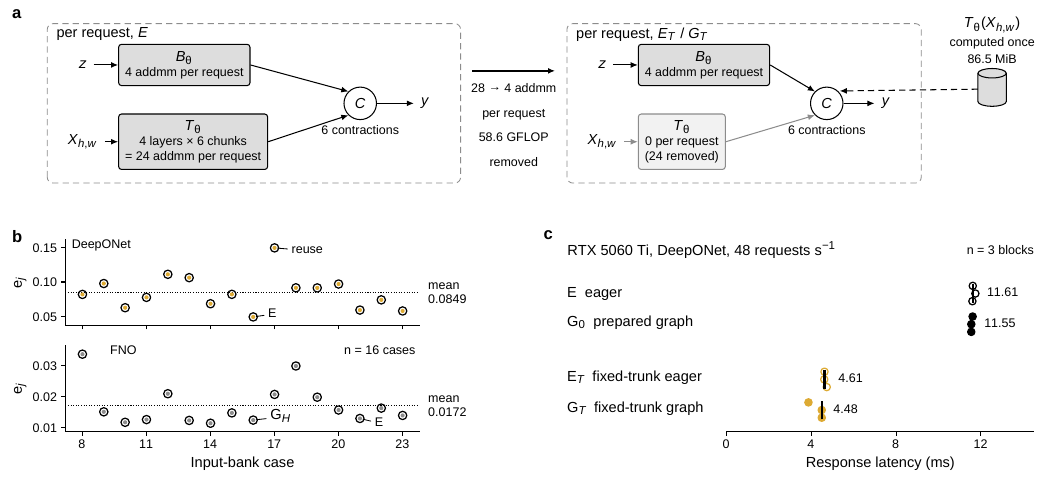}
\caption{\textbf{Fixed-trunk reuse removes repeated DeepONet work.}
\textbf{a}, Retaining four trunk layers over six query chunks removes 58.6\,GFLOP per request for 86.5\,MiB of storage. Dense calls fall from 28 to four; six contractions remain unchanged.
\textbf{b}, Decoded relative $L_2$ errors against truth for 16 input-bank cases per model. Reference and candidate markers coincide; $e_j$ denotes the error for input-bank case $j$. Models use different native grids; the complete audit contains 80 comparisons (Supplementary Table~\ref{tab:si-darcy-physical}).
\textbf{c}, Response latency for the RTX~5060~Ti DeepONet blocks in Fig.~\ref{fig:service}a. Points are block medians ($n=3$ per arm); bars mark arm medians.}
\label{fig:physical}
\end{figure}
\clearpage
\begin{figure}[t]
\centering
\includegraphics[width=\linewidth]{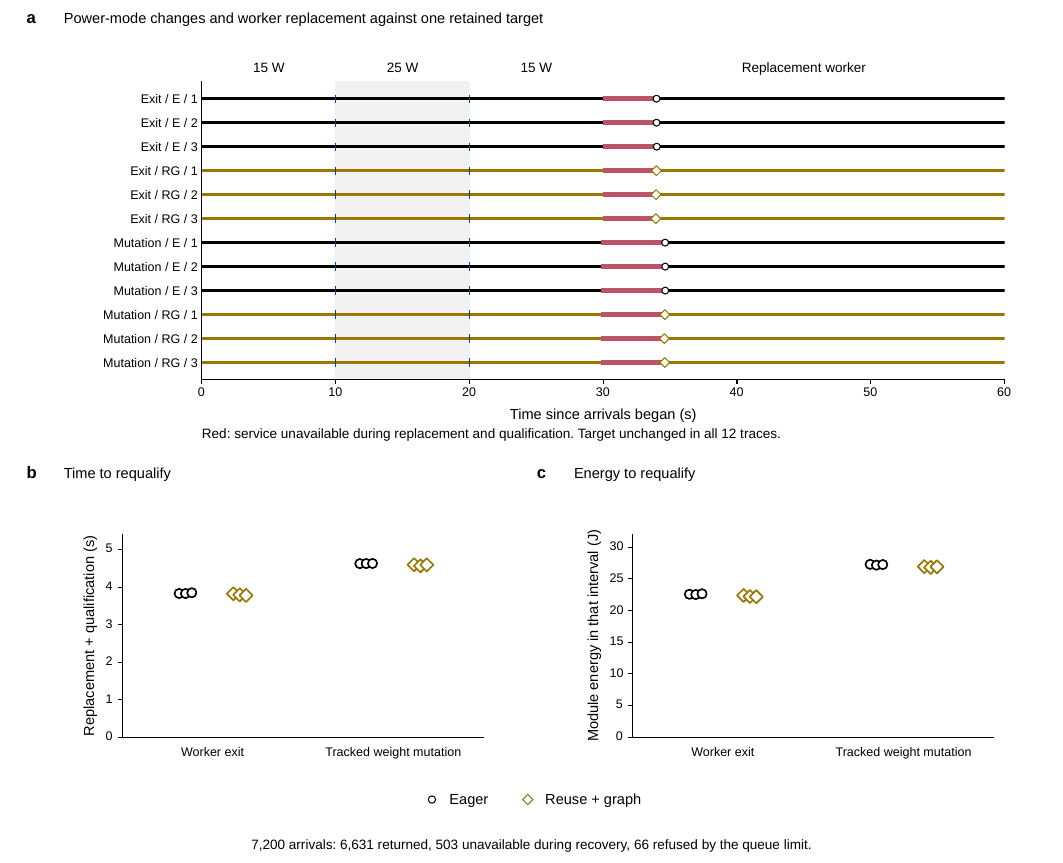}
\caption{\textbf{Rebuilding a worker preserves the target but interrupts service.}
\textbf{a}, Twelve traces ($n=3$ per arm and fault). Power-mode changes at 10 and 20\,s retain the worker. A weight mutation or exit at 30\,s triggers replacement; red intervals end when the new worker qualifies against the original target. Events occur between requests.
\textbf{b,c}, Replacement time and module energy, including controller instrumentation. Each point is one trace. Replacement energy is part of the trace total. All returned outputs match the target; unavailable arrivals and queue refusals are counted separately in Supplementary Note~\ref{si:edge-service}.}
\label{fig:edge_recovery}
\end{figure}
\clearpage
\endgroup

%% file: si/05_si.tex
\setcounter{table}{0}
\emergencystretch=2em
\providecommand{\OpCertMainRef}[1]{\ref{#1}}
\providecommand{\OpCertSInote}[2]{%
  \FloatBarrier
  \refstepcounter{subsection}%
  \subsection*{Supplementary Note~\thesubsection. #2}\label{#1}}
\setcounter{subsection}{0}
\renewcommand{\thesubsection}{\arabic{subsection}}
\setcounter{figure}{0}
\renewcommand{\thefigure}{\arabic{figure}}
\providecommand{\theHfigure}{}
\renewcommand{\theHfigure}{opcert.si.figure.\arabic{figure}}
\captionsetup[figure]{name=Supplementary Fig.}
\renewcommand{\thetable}{\arabic{table}}
\setcounter{equation}{0}
\renewcommand{\theequation}{S\arabic{equation}}
\providecommand{\theHtable}{}
\renewcommand{\theHtable}{opcert.si.table.\arabic{table}}
\providecommand{\theHequation}{}
\renewcommand{\theHequation}{opcert.si.equation.\arabic{equation}}
\providecommand{\theHsubsection}{}
\renewcommand{\theHsubsection}{opcert.si.note.\arabic{subsection}}
\captionsetup[table]{name=Supplementary Table}
\ifdefined\OpCertSIStandalone\else
  \section*{Supplementary Information}
\fi

\ifdefined\OpCertSIStandalone
These notes give the protocols, numerical audits and controls supporting the main results. The overview identifies each cohort and its measurement boundary.
\fi

\input{si/cohort_map}

\input{si/source_to_edge}

\input{si/edge_service}

\OpCertSInote{si:service}{Response latency and request energy}

This note supports Fig.~\OpCertMainRef{fig:service}, comparing five DeepONet and three FNO execution paths within the same CPU-to-CPU service. The checkpoints were trained using relative $L_2$ loss and selected by validation relative $L_2$. Each is retained to isolate execution costs; the different native grids prevent an accuracy or efficiency ranking between the two model families. Prepared-input configuration controls appear separately in Note~\ref{si:legacy}.

\subsubsection*{Requests and measurement interval}
Positions 0--7 of each 24-input bank qualify the worker; positions 8--23 are served. Before timing, each path evaluates the serving inputs twice against their retained targets, so these are distinct but not unseen validation inputs. Each fresh worker warms for at least five seconds and one complete input cycle. Construction failure is recorded separately from numerical disagreement. All 72 blocks completed:
24 per device, with 576, 96 and 240 requests per block on RTX~5060~Ti,
Jetson and A2000, respectively. Each block uses a 12~s arrival horizon and
at least 5~s warmup outside the energy window. Rates are therefore 48, 8
and 20~Hz for both models. Three fixed execution orders are used; five-arm
DeepONet is not completely position-balanced, and odd-group pairwise
precedence is not balanced. These are within-device paced comparisons,
not a saturation benchmark or a cross-device rate-controlled ranking.

Each request encodes the resident normalised CPU tensor and transfers its shape, dtype and complete data to a spawned worker. The worker decodes into owned CPU storage, copies into a persistent GPU input buffer, executes and synchronises, and returns a complete independent CPU output. Only one request is in flight; the controller verifies request, worker-generation and reference identities. Previously returned fields therefore remain valid across buffer reuse. Dispatch latency spans input encoding to receipt of the complete encoded CPU response; response time additionally includes scheduled-arrival delay. Input acquisition and scientific preprocessing precede this boundary, while physical decoding follows it.

\subsubsection*{Costs and paired comparisons}
Table~\ref{tab:si-control-costs} reports every arm, including the empirical
comparator $G_H$. Main Fig.~\OpCertMainRef{fig:service} draws the 54 blocks
of the other six arms. Table~\ref{tab:si-control-contrasts}
separates replay, explicit reuse and empirical-pass changes. For group $g$,
\begin{equation}
 e_{a,g}=\frac{E_{a,g}}{N^{\mathrm{completed}}_{a,g}},
 \qquad
 \delta_{a\to b,g}=1-\frac{m_{b,g}}{m_{a,g}},
 \label{eq:si-control-contrast}
\end{equation}
where $E$ is interval energy, $N^{\mathrm{completed}}$ is the completed-request count, $e$ is energy per request, $a$ and $b$ denote execution arms, and $m$ is response latency or request energy. Three group-matched contrasts are summarised by their median and observed minimum-to-maximum range. The contrasts are not assumed additive. Here response latency is the block median; energy is the integrated interval energy divided by completed requests. The ratio of arm medians need not equal the median paired ratio.

\begin{table}[!htbp]
\centering\footnotesize
\caption{\textbf{All execution paths in the paced-service cohort.} Entries are medians over three blocks. E, qualified eager; $G_0$, prepared DeepONet or native FNO graph; $E_T$, explicit trunk eager; $G_T$, explicit trunk graph; $G_H$, empirical-pass graph. Memory is the qualification snapshot, not a peak. Energy is GPU-total on desktops and module-input on Jetson. Preparation ends at readiness and is outside request energy.}
\label{tab:si-control-costs}
\setlength{\tabcolsep}{4pt}
\begin{tabular}{@{}llrrrrr@{}}
\toprule
Device / arm & Rate & Response & Energy & Prep. & Alloc. & Reserv. \\
 & (Hz) & (ms) & (J/request) & (s) & (MiB) & (MiB) \\
\midrule
\multicolumn{7}{l}{\textbf{DeepONet}} \\
5060 Ti E & 48 & 11.615 & 1.817 & 1.458 & 101.5 & 222.0 \\
5060 Ti $G_0$ & 48 & 11.545 & 1.811 & 1.488 & 166.2 & 308.0 \\
5060 Ti $E_T$ & 48 & 4.613 & 0.482 & 1.353 & 188.1 & 254.0 \\
5060 Ti $G_T$ & 48 & 4.484 & 0.480 & 1.336 & 252.7 & 286.0 \\
5060 Ti $G_H$ & 48 & 11.576 & 1.818 & 1.551 & 166.2 & 308.0 \\
\addlinespace
Jetson E & 8 & 76.649 & 2.012 & 4.279 & 101.5 & 222.0 \\
Jetson $G_0$ & 8 & 75.950 & 2.007 & 4.580 & 166.2 & 310.0 \\
Jetson $E_T$ & 8 & 16.503 & 1.038 & 3.369 & 188.1 & 254.0 \\
Jetson $G_T$ & 8 & 15.633 & 1.038 & 3.476 & 252.7 & 288.0 \\
Jetson $G_H$ & 8 & 76.353 & 2.003 & 5.142 & 166.2 & 310.0 \\
\addlinespace
A2000 E & 20 & 39.854 & 2.686 & 4.381 & 101.5 & 222.0 \\
A2000 $G_0$ & 20 & 39.268 & 2.801 & 4.650 & 166.2 & 308.0 \\
A2000 $E_T$ & 20 & 15.445 & 1.654 & 4.396 & 188.1 & 254.0 \\
A2000 $G_T$ & 20 & 16.098 & 1.652 & 4.230 & 252.7 & 286.0 \\
A2000 $G_H$ & 20 & 39.378 & 2.600 & 4.790 & 166.2 & 308.0 \\
\addlinespace
\multicolumn{7}{l}{\textbf{FNO}} \\
5060 Ti E & 48 & 2.263 & 0.524 & 1.251 & 57.1 & 74.0 \\
5060 Ti $G_0$ & 48 & 1.621 & 0.513 & 1.254 & 121.2 & 160.0 \\
5060 Ti $G_H$ & 48 & 1.743 & 0.511 & 1.270 & 171.5 & 200.0 \\
\addlinespace
Jetson E & 8 & 10.537 & 1.047 & 3.065 & 57.1 & 74.0 \\
Jetson $G_0$ & 8 & 8.103 & 1.043 & 3.134 & 121.2 & 162.0 \\
Jetson $G_H$ & 8 & 8.726 & 1.043 & 3.328 & 171.5 & 202.0 \\
\addlinespace
A2000 E & 20 & 7.779 & 1.653 & 3.776 & 57.1 & 74.0 \\
A2000 $G_0$ & 20 & 6.289 & 1.642 & 3.915 & 121.2 & 160.0 \\
A2000 $G_H$ & 20 & 6.266 & 1.636 & 3.861 & 171.5 & 200.0 \\
\addlinespace
\bottomrule
\end{tabular}
\end{table}

\begin{table}[!htbp]
\centering\footnotesize
\caption{\textbf{Matched contrasts separate execution mechanisms.} Each cell gives median percentage reduction and observed range over three paired groups. Negative values indicate increases. Each contrast uses the same request trace and service boundary.}
\label{tab:si-control-contrasts}
\setlength{\tabcolsep}{4pt}
\begin{tabular}{@{}llrr@{}}
\toprule
Device & Contrast & Response reduction (\%) & Energy reduction (\%) \\
\midrule
\multicolumn{4}{l}{\textbf{DeepONet}} \\
5060 Ti & $E\to G_0$ & 0.46 [0.06, 1.64] & 0.19 [-0.17, 0.33] \\
5060 Ti & $E\to E_T$ & 60.28 [59.31, 60.79] & 73.42 [73.36, 73.87] \\
5060 Ti & $G_0\to G_T$ & 61.16 [61.16, 66.78] & 73.49 [73.25, 73.69] \\
5060 Ti & $E_T\to G_T$ & 4.99 [2.58, 16.41] & -0.52 [-0.58, 0.59] \\
5060 Ti & $G_0\to G_H$ & -0.18 [-1.10, 0.28] & -0.15 [-0.67, -0.13] \\
\addlinespace
Jetson & $E\to G_0$ & 0.93 [0.91, 0.94] & 0.37 [0.24, 1.27] \\
Jetson & $E\to E_T$ & 78.42 [78.37, 78.82] & 48.43 [48.43, 49.65] \\
Jetson & $G_0\to G_T$ & 79.37 [79.27, 79.54] & 48.36 [48.30, 48.93] \\
Jetson & $E_T\to G_T$ & 5.27 [3.04, 6.24] & -0.01 [-0.15, 0.23] \\
Jetson & $G_0\to G_H$ & -0.49 [-0.53, -0.46] & 0.22 [-0.52, 0.31] \\
\addlinespace
A2000 & $E\to G_0$ & 1.47 [0.86, 5.53] & -2.08 [-4.28, 2.01] \\
A2000 & $E\to E_T$ & 60.93 [60.07, 61.25] & 40.63 [38.08, 43.42] \\
A2000 & $G_0\to G_T$ & 58.83 [58.65, 61.07] & 41.84 [41.02, 42.12] \\
A2000 & $E_T\to G_T$ & 1.03 [-4.46, 2.17] & 0.01 [-0.25, 0.68] \\
A2000 & $G_0\to G_H$ & -0.23 [-0.93, 0.20] & 7.23 [-1.34, 9.22] \\
\addlinespace
\multicolumn{4}{l}{\textbf{FNO}} \\
5060 Ti & $E\to G_0$ & 28.89 [28.40, 30.04] & 2.66 [1.55, 3.08] \\
5060 Ti & $G_0\to G_H$ & -7.35 [-10.83, -7.09] & 0.21 [0.05, 0.63] \\
\addlinespace
Jetson & $E\to G_0$ & 23.56 [23.10, 24.92] & -0.05 [-0.26, 0.35] \\
Jetson & $G_0\to G_H$ & -7.70 [-8.09, -7.43] & 0.10 [-0.03, 0.76] \\
\addlinespace
A2000 & $E\to G_0$ & 20.42 [13.62, 24.56] & 0.35 [-1.13, 0.64] \\
A2000 & $G_0\to G_H$ & 4.71 [-8.68, 8.85] & 0.46 [-0.48, 1.61] \\
\addlinespace
\bottomrule
\end{tabular}
\end{table}

\subsubsection*{Output comparisons and energy accounting}
All 21,888 returned tensors passed the recorded post-block shape, dtype and
byte comparisons after the energy interval, with completed outputs held in client memory until then. Incomplete blocks, mismatches and unusable telemetry stop the campaign sequence and remain archived. The independent saved-array audit covered all 1,152 full
exemplars, 16 per block, together with their supplied inputs. The remaining
20,736 repeated outputs retain the original checks and digests but not full
arrays for an independent later comparison. Separately audited gate and
history outputs are not added to the performance sample count.

All 48 discrete-GPU blocks used cumulative NVML energy counters; all 24 Jetson blocks used integrated module-input power. The 20~ms host polling interval describes software polling rather than sensor refresh. The polled power traces are diagnostic: on RTX devices their 12~s means differ from the cumulative-counter means by up to 4.3~W, and request energy uses the counters. Counter endpoints and host brackets support the interval audit. GPU-total and module-input domains include idle intervals and observed co-use without subtraction or per-process attribution \citep{nvidia2026nvml,nvidia2026jetsonpower}. Because co-use was observed rather than experimentally fixed, small paired energy differences are not attributed solely to execution strategy. Numerical comparison, physical decoding, preparation and warmup are outside the request-energy interval. The interval begins before the first timed input encoding and ends after delivery, spanning at least the full arrival horizon.

Construction-to-readiness time and qualification-memory snapshots are reported for every arm. The timings include each strategy's construction and qualification schedule. Allocated and reserved storage are measured at readiness, rather than over the worker lifetime. Reference generation, failed admission, retirement and recovery energy are outside the request-energy blocks.

An earlier 45-block cohort used a different arm set and offered rates. Its complete records, including preparation and interrupted attempts, remain in the source archive; they are not additional replicates of the 72-block comparison.

\OpCertSInote{si:profiling}{Computation and service-path profiling}

To distinguish reduced GPU computation from transport costs, we profile the DeepONet paths in Fig.~\OpCertMainRef{fig:service}. The instrumented diagnostic contains one Nsight Systems 2026.1.3 capture for each
of eager, graph and reuse, with 16 measured requests per arm after setup and
16 warmup requests. The worker executes CUDA activity inside a parent-marked
measurement interval. Table~\ref{tab:si-service-trace} retains that process
and interval distinction. Profiled timings are diagnostic observations,
separate from the three-group uninstrumented service estimates.

\begin{table}[htbp]
\centering\small
\caption{\textbf{Instrumented service trace over 16 measured requests on RTX 5060 Ti.}
The parent interval contains client/service activity. GPU entries are sums
of individually traced kernel durations for eager, or graph execution
intervals for graph and reuse. The latter do not expose individual kernels
inside replay. These overlapping timing categories are not additive.}
\label{tab:si-service-trace}
\setlength{\tabcolsep}{4pt}
\begin{tabular}{@{}p{0.49\linewidth}rrr@{}}
\toprule
Quantity & Eager & Graph & Reuse \\
\midrule
Parent measured interval (ms) & 186.276 & 184.113 & 57.781 \\
GPU kernel-duration sum (ms) & 137.698 & Not resolved & Not resolved \\
GPU graph-interval sum (ms) & n/a & 138.294 & 6.806 \\
CUDA graph launches & 0 & 16 & 16 \\
Host-to-device bytes & 34030336 & 34030272 & 34030272 \\
Device-to-host bytes & 11343424 & 11343424 & 11343424 \\
\bottomrule
\end{tabular}
\end{table}

Graph and reuse each launch one graph per request. Their GPU graph-interval sums differ by a factor of 20.32, while their parent intervals differ by 3.19; both retain 2,835,856 transferred bytes per request. The equal transfer volume and shorter reuse graph intervals support removal of repeated GPU computation while retaining transport and host-service work. Kernel durations, graph spans and CPU synchronisation overlap and are not used as an exclusive latency decomposition. Graph-level collection resolves replay intervals but not their individual kernels.

Trace selection uses the completed parent-process NVTX interval that
contains the 16 measured requests. String-table identifiers are resolved
to recover the interval name. Worker CUDA events are selected when both
their start and end timestamps lie within that interval, using the common
trace clock rather than a parent-only process association. Eager entries
sum traced kernel durations; graph entries sum replay intervals. This
cross-process selection recovers GPU work without assigning overlapping
wait, copy and execution intervals to disjoint costs. The exported traces
and selection procedure are retained in the archived experimental records.

\OpCertSInote{si:physical}{Physical decoding and prediction error}

This note tests whether the execution changes preserve decoded fields and their existing prediction errors (Extended Data Fig.~\OpCertMainRef{fig:physical}b). The Darcy audit covers checked eager, empirical-pass graph $G_H$ and DeepONet fixed-trunk reuse. The timed $G_0$, $E_T$ and $G_T$ arms were not decoded separately. The audit applies the unchanged checkpoint-native inverse normalisation
on CPU to retained normalised fields, then scores the decoded prediction
against the existing simulated solution on each model's native grid. Shape,
dtype and complete bytes are compared both before and after decoding.
Normalised equality remains required even if decoding could collapse a
numerical difference. Physical relative $L_2$ is the norm of decoded
prediction minus truth divided by the truth norm, using float64 reductions
without a denominator epsilon. No interpolation between model grids is used.

\begin{table}[htbp]
\centering\small
\caption{\textbf{Darcy outputs before and after physical decoding.} Three DeepONet paths and two FNO paths each use positions 8--23 on RTX~5060~Ti, giving 80 comparisons. Counts denote complete-output agreement with each reference; relative $L_2$ is the mean error against truth on the native grid. Graph denotes $G_H$ in this separate audit.}
\label{tab:si-darcy-physical}
\setlength{\tabcolsep}{4pt}
\begin{tabular}{@{}llrrr@{}}
\toprule
Model and native grid & Candidate & Normalised exact & Decoded exact & Mean relative $L_2$ \\
\midrule
DeepONet, $421\times421$ & Checked eager & 16/16 & 16/16 & 0.0848994 \\
 & Graph & 16/16 & 16/16 & 0.0848994 \\
 & Reuse & 16/16 & 16/16 & 0.0848994 \\
FNO, $85\times85$ & Checked eager & 16/16 & 16/16 & 0.0172171 \\
 & Graph & 16/16 & 16/16 & 0.0172171 \\
\bottomrule
\end{tabular}
\end{table}

These 80 comparisons supply the physical-error panel in Extended Data Fig.~\OpCertMainRef{fig:physical}b. Decoding, truth loading and scoring occur outside the energy interval. They preserve each checkpoint's existing error on its native grid; the different resolutions prevent a controlled accuracy ranking between models.

The physical audit uses the simulated solution field from the existing Darcy smooth2 test data. Checkpoint-native inverse normalisation and target loading are unchanged, and the archived provenance records associate each checkpoint with its retained reference and per-arm outputs. Neither the decoder nor the truth arrays were selected by candidate error. Compact figure exports reproduce the displayed fields and per-position scores; reproducing the full 80-comparison audit additionally requires the retained raw outputs and cohort decoder.

\subsubsection*{Accuracy limits of the shear-flow history control}
The Well shear-flow case serves a different purpose. Four input windows were reconstructed from shear-flow trajectory zero on its native $256\times512$ grid. Windows 20 and 40 qualify the worker; windows 60 and 80 are the two serving cases, with next-state truth at steps 64 and 84. Under the recovered CPU inverse, all 42 returned fields in the resource-history service cohort match the normalised reference and decoded output. A fixed diagnostic compares the model, persistence and native-zero predictions for every channel of all four windows. On the two served windows, model/persistence mean-squared-error ratios are 9.27--34.43 for tracer, $4.11\times10^4$--$2.15\times10^5$ for pressure, 455.96--1,644.68 for velocity component 0 and 37.71--143.43 for component 1. The complete channel/window scores remain in the source records.

Every served channel is worse than persistence under this decoder. Small pressure truth RMS amplifies its relative $L_2$, and demeaning does not reverse the baseline comparison. The available artifacts do not fully resolve the checkpoint's training/export contract, including the run-specific direct-state versus delta-target choice. The decoder was not selected by error minimisation. We therefore use this checkpoint for numerical-history tests, with the fixed-window physical scores retained as a control on its interpretation.

\OpCertSInote{si:legacy}{Operator coverage and prepared-input reuse costs}

Supporting cohorts provide broader admission, recovery and deployment evidence, while the paced-service cohort provides the CPU-input-to-CPU-output cost comparison. Each cohort retains its recorded software, numerical settings, comparison predicate and measurement interval and is analysed separately from the paced-service estimates.

\subsubsection*{Admission across operator families}
The operator-admission cohort examined 42 checkpoint/configuration cells (Table~\ref{tab:admission-outcomes}). Its logical-bit predicate included output structure, tensor shape, dtype, device kind and logical element bits. Candidate matching and reference repeatability use the retained 64-bit digests and scalar differences; complete arrays are unavailable for independent comparison of every instance. Selection among passing candidates uses their measured operating costs.
Unavailable candidates and cells without a repeatable reference remain admission outcomes. On Jetson, deterministic/default eager latency ratios for the six initially nonrepeatable pairs range from 0.99 to 8.56 (median 1.55); the full pair-level record is archived with the admission data. The 29 selected graphs comprise 20 of 32 Jetson cells and nine of ten RTX~5060~Ti cells; their median fractional latency reductions at the prepared-input boundary were 0.75 and 0.37, respectively. Table~\ref{tab:admission-outcomes} retains every cell, including nonrepeatable references and unsuccessful constructions.

\begin{table}[!htbp]
\centering\small
\caption{\textbf{Complete reference, hoisting and graph outcomes for the 42 configuration cells.} Original and deterministic configurations of a checkpoint are separate cells, not independent applications. P: eager repeated and both hoisting and graph capture agreed. F: eager repeated and hoisting agreed, but graph construction failed. U: eager repeated and hoisting agreed, but graph was unavailable. N: eager did not repeat, so candidates were not qualified. A dash denotes an untested checkpoint/configuration.}
\label{tab:admission-outcomes}
\setlength{\tabcolsep}{5pt}
\begin{tabular}{@{}lcccc@{}}\toprule
 & \multicolumn{2}{c}{Jetson} & \multicolumn{2}{c}{RTX 5060 Ti} \\
Checkpoint / task & Original & Deterministic & Original & Deterministic \\\midrule
DeepONet, Burgers $r512$ & P & P & -- & -- \\
DeepONet, Darcy $r421$ & P & P & -- & -- \\
FNO, Burgers $r512$ & P & P & -- & -- \\
FNO, Darcy $r85$ & P & P & -- & -- \\
NEUTRON, lid-driven cavity & N & P & P & P \\
NEUTRON, plasticity & N & P & -- & -- \\
Sp2GNO, Burgers $r512$ & N & F & -- & -- \\
Sp2GNO, Burgers $s2048$ & N & F & -- & -- \\
Sp2GNO, Darcy $r141$ & N & F & N & P \\
WNO, Burgers $r2048$ & P & P & -- & -- \\
WNO, Darcy $r141$ & N & P & -- & -- \\
DPOT-L & U & U & P & P \\
DPOT-M & P & P & -- & -- \\
MoE-POT small & P & P & -- & -- \\
MoE-POT tiny & P & P & P & P \\
Poseidon-B & P & P & P & P \\
\bottomrule\end{tabular}
\end{table}

Jetson contributes 32 cells and RTX~5060~Ti ten; 26 and nine, respectively, established repeatable references. All 35 hoisted candidates agreed, as did 30 graph candidates; three graph constructions failed and two were unavailable. Compilation was attempted in the nine reference-qualified workstation cells only, using default and reduce-overhead modes. All nine default-mode and seven constructed reduce-overhead candidates differed from eager. The other two reduce-overhead candidates, both NEUTRON lid-driven-cavity configurations, were unavailable. Thus the 88 candidate records comprise 65 agreeing, 16 differing, three failed and four unavailable outcomes. No compiler candidates were attempted on Jetson.

Allocator and recovery protocols are consolidated in Supplementary Note~\ref{si:history}.

In a separately profiled deterministic Poseidon-B case, both compiler modes repeated and agreed with each other, but 60,376 of 65,536 elements differed from eager, with maximum absolute difference $2.146\times10^{-6}$. Repeatable optimised arithmetic can therefore differ from the chosen reference; the experiment does not evaluate the physical acceptability of that difference. Original repeat-evaluation records for the additional self-trained checkpoints, including the failed allocator evaluation, remain archived.

\subsubsection*{Supervision and prepared-input device comparisons}
The prepared-input supervision cohort measures prepared device input through owned device output using nine qualification and 15 evaluation positions from each 24-input bank. In its matched graph control, three fresh processes per arm on each of RTX~5060~Ti, Jetson and RTX~A2000 give 18 processes in total. The graph and input/output ownership are held fixed while supervision is added. Independent before/after comparisons matched all 864 saved fields, with median paired latency increases of 1.16\%, 1.37\% and 1.19\%, respectively. No energy block was recorded for this control, and numerical comparison was not performed online for every timed output. These measurements quantify supervision at the prepared-device boundary.

The companion A40, A100, GH200 and H200 replication ran 22 fresh processes per device, with 1,200 saved full-field comparisons matching the device-local reference on each device. These observations support portability of the prepared-input execution components and reference checks across those devices. Because the interface differs from the paced CPU-to-CPU service, the results are reported separately. The saturated energy blocks exclude setup, warmup and host energy, including the Grace CPU on GH200, and therefore remain component-level measurements rather than a seven-device paced-service estimate.

Table~\ref{tab:seven-device-costs} retains the corresponding prepared-input latency and saturated-energy measurements.

\begin{table}[!htbp]
\centering\small
\caption{\textbf{Prepared-input Darcy DeepONet costs on seven devices.} Medians of three fresh-process blocks. Latency spans device input to owned device output. Energy uses separate saturated intervals without setup or warmup; eager energy was not recorded. Discrete devices report GPU energy and Jetson reports module-input energy. Compare paths within each device.}
\label{tab:seven-device-costs}
\setlength{\tabcolsep}{5pt}
\begin{tabular}{@{}lrrrrr@{}}\toprule
 & \multicolumn{3}{c}{Latency (ms)} & \multicolumn{2}{c}{Energy (J/request)} \\
Device & Eager & Graph & Reuse & Graph & Reuse \\\midrule
H200 & 2.642 & 2.188 & 0.430 & 1.1375 & 0.0850 \\
GH200 & 4.911 & 2.392 & 0.566 & 1.1362 & 0.1101 \\
A100 & 6.941 & 6.169 & 0.727 & 1.9550 & 0.1115 \\
A40 & 8.882 & 8.232 & 0.857 & 2.4042 & 0.1192 \\
RTX 5060 Ti & 10.894 & 10.700 & 0.693 & 1.8769 & 0.0606 \\
RTX A2000 & 29.734 & 29.565 & 1.797 & 1.9959 & 0.1237 \\
Jetson Orin Nano & 67.874 & 65.542 & 3.243 & 1.5034 & 0.0587 \\
\bottomrule\end{tabular}
\end{table}

\subsubsection*{Resolution and reuse mechanism}
A resolution study evaluated ten existing DeepONet checkpoints, five each for Burgers and Darcy, without retraining for deployment. The Burgers endpoints span 512 to 8,192 queries, with paired supervised-graph/reuse latency ratios of 1.20 to 2.83; Table~\ref{tab:reuse-resolution} gives all five Darcy configurations. Four execution strategies ran in three fresh processes per checkpoint, giving 120 processes, each with 128 timed requests after ten warmups. This cohort uses its own deterministic configuration, nine qualification positions and 15 evaluation positions. All 5,760 saved full fields, 24 before and 24 after timing per process, matched the local references; no energy was measured.

The Darcy subset compares the supervised graph with supervised fixed-trunk graph at five configurations on RTX~5060~Ti. CPU preprocessing, host-to-device input transfer, physical decoding and IPC are excluded. Both checkpoint and sensor resolution vary, so these are configuration-dependent reuse benefits rather than an isolated effect of query-grid size.

\begin{table}[!htbp]
\centering\small
\caption{\textbf{Reuse relative to a graph-based execution with the same supervisor.} Latencies are arm medians; speedup is the median of three paired process ratios with observed range. This is a prepared-device-input cohort, not the full service.}
\label{tab:reuse-resolution}
\begin{tabular}{@{}rrrr@{}}\toprule
Grid & Graph (ms) & Reuse + graph (ms) & Paired speedup \\\midrule
$85^2$ & 0.5661 & 0.2587 & 2.209 (2.164--2.266) \\
$141^2$ & 1.2585 & 0.2753 & 4.565 (4.550--4.647) \\
$211^2$ & 2.7645 & 0.3245 & 8.518 (8.421--8.656) \\
$281^2$ & 4.8058 & 0.4656 & 10.327 (10.322--10.725) \\
$421^2$ & 10.7004 & 0.6929 & 15.435 (15.431--15.561) \\
\bottomrule\end{tabular}
\end{table}

The checkpoint audit verifies 40 bundle-manifest file hashes and all 60 original Darcy four-arm result hashes in this configuration campaign. It independently compares 1,440 saved supervisor-arm output arrays with their references; all agree. This audit reloads the complete saved arrays, whereas the sustained-service audit checks the archived flags and executed comparison code.

Removed dense work ranged from 0.169 to 58.6\,GFLOP per request. An effective-throughput fit to saved time had 5.3\% median absolute relative error in the fitted speed-up values and 18.2\% maximum error. The fit describes these checkpoints on the measured device. Supervision added 0.121--0.132\,ms at the prepared-input boundary.

Instrumented native and fixed-trunk eager traces provide a dependency-level explanation of DeepONet reuse. Three requests per arm recorded 84 versus 12 \texttt{addmm} calls, or 28 versus four per request. Both arms recorded 18 \texttt{einsum} and 18 \texttt{bmm} calls: six logical contractions per request, with \texttt{bmm} implementing them rather than adding six further contractions. These are instrumented eager observations rather than host-operation counts during graph replay. The removed calls were four trunk layers over six original chunks; their recorded shapes account for 58,622,815,232 dense-operation FLOPs per request, counting a multiply-add as two operations and excluding bias, activation and coordinate construction. The retained trunk tables occupied 90,747,392 bytes. This quantifies the work-storage exchange in fixed-grid trunk precomputation \citep{winovich2025active}. The cited method also restructures batched output evaluation and is not used as a batch-one GPU-service timing baseline. The profiling in Supplementary Note~\ref{si:profiling} supplies the separate GPU-work observation. The operation counts explain the removed work for this checkpoint.

For the scalar-output Darcy model, a float32 trunk table with $N_q$ queries and width $p$ requires
$M_T=4N_qp$ bytes (86.5\,MiB for $N_q=421^2$, $p=128$). Total worker memory
also includes activations and workspace. Reading each table element once
for a dot product gives a table-read-only intensity of
$2N_qp/(4N_qp)=0.5$ FLOP per byte, excluding cache reuse and other traffic.
This is an analytical estimate, not a measured bandwidth limit. For $q$ separately retained output-channel representations, this storage becomes $4N_qpq$ bytes, as in the four-channel heat-exchanger model. Query count and worker multiplicity constrain full-table reuse.

\OpCertSInote{si:scope}{Numerical references and comparison scope}

A retained numerical reference is interpreted within the checkpoint, input specification, declared numerical configuration and output boundary with which it was generated. References in the Darcy and supporting cohorts are device- and configuration-local. The continuous-inference MIMONet comparison in Note~\ref{si:source-edge} retains a common workstation target, against which all measured Jetson executions fail byte identity. The service in Note~\ref{si:edge-service} uses the archived Jetson-local eager target. Prediction accuracy, repeatability, retained-reference agreement and request completion are reported separately. The qualification record is defined in the main-text Methods. Device-specific numerical settings in this note apply to the Darcy cohort; the deterministic heat-exchanger configuration is reported separately in Methods.

For outputs $y$ and retained fields $r$, byte-identity agreement is defined as follows:
\begin{equation}
E_{\mathrm{bit}}(y,r)=1
\quad\Longleftrightarrow\quad
\begin{aligned}
&\operatorname{finite}(y)\land\operatorname{finite}(r),\\
&\operatorname{structure}(y)=\operatorname{structure}(r),\\
&\operatorname{shape}(y)=\operatorname{shape}(r),\qquad
\operatorname{dtype}(y)=\operatorname{dtype}(r),\\
&\operatorname{bits}_{\mathrm{logical}}(y)=\operatorname{bits}_{\mathrm{logical}}(r).
\end{aligned}
\label{eq:si-output-contract}
\end{equation}
For structured or multi-tensor outputs, the conditions apply componentwise. No numerical tolerance is used. Signed zeros remain distinguishable. Finiteness is required before comparison; physical validity requires separate evidence. Storage address, stride and offset are not part of numerical equality. Device placement belongs to the declared interface, which returns a complete CPU tensor in the paced-service cohort. Cohort-specific predicates and interfaces are identified in Supplementary Notes~\ref{si:legacy} and~\ref{si:contractfrontier}.

Each numerical comparison is recorded with its checkpoint, input, reference, candidate, execution environment and observation boundary and is identified as a qualification witness, delivered request or retained array. Performance blocks are separate measurement units. Repeating one input across requests, arms, blocks or devices adds execution observations rather than new scientific test inputs.

For the paced-service cohort, expected serving answers remain with the experimental parent rather than the worker. Supplementary Note~\ref{si:service} gives the input partition, prechecks, post-block comparisons and independent saved-array audit. Earlier A2000 smoke runs remain separate from those estimates.

\subsubsection*{Checkpoint, input and environment provenance}
The Darcy cohort uses the retained seed-zero grid-based DeepONet and two-dimensional FNO checkpoints on their native $421\times421$ and $85\times85$ grids. Positions 0--7 qualify the execution and positions 8--23 are served. The corresponding input tensors have shapes $1\times421\times421\times3$ and $1\times85\times85\times3$, and the service returns complete normalised scalar fields. The provenance records identify the checkpoint, serialised model, input bank and workload for each experiment.

The retained reference files do not embed checkpoint identity. During staging, each reference is associated with its bundled checkpoint, and subsequent checks verify that association. The archived source identities and protocol associations reproduce this staging step; they record the numerical target rather than establish its application accuracy.

The paced Darcy service uses PyTorch \texttt{2.13.0+cu129} on RTX~5060~Ti and A2000, and \texttt{2.5.0a0+872d972e41.nv24.08} on Jetson. Desktop energy covers the GPU; Jetson energy covers module input. Driver, operating-system and power-mode identifiers were not retained for this cohort.

All six inspected qualification records specify float32, one CPU thread,
enabled cuDNN, disabled cuDNN benchmarking, disabled deterministic
convolution and deterministic-algorithm enforcement, allowed cuDNN TF32,
disallowed matrix-multiplication TF32, \texttt{highest} float32 matrix
precision, disabled autocast and \texttt{CUBLAS\_WORKSPACE\_CONFIG=:4096:8}.
Admission uses a 1,536~MiB allocator ceiling. Each reference retains these device-specific conditions. Unrecorded environment fields remain unspecified.
\OpCertSInote{si:qualification}{Qualification and execution controls}
\label{sec:reference-proof}

This note gives the runtime checks and the additional conditions needed to preserve outputs over an input domain. Qualification compares complete output bytes with a retained reference before service; live guards inspect exposed settings, state and resource conditions. Note~\ref{si:service} specifies the input/output and storage-ownership boundaries.

In a DeepONet control on RTX~5060~Ti, one bit of the serialised output was flipped after the runtime checks. The eager, graph and reuse paths returned the altered output, which the external byte comparison detected. The following argument states the additional conditions needed for preservation over a declared input domain.

For a checkpoint $\theta$, input $x$ and numerical configuration $\sigma$, execution depends on worker state $s$ and may update it to $s'$:
\begin{equation}
 (y,s')=F_{\theta}(x;\sigma,s),
 \label{eq:si-stateful-execution}
\end{equation}
Here $y$ is the output, $s'$ is the state after execution, and $s$ includes allocator and execution-plan state. The reference corpus $\mathcal R$ contains qualification input--output pairs $\{(w_j,r_j)\}$ from designated runs; it does not serialise the execution state for every possible input. A sufficient preservation argument requires four conditions: (H1) corresponding arithmetic uses the same deterministic realisation and operation order; (H2) reused values are produced by the reference path from immutable dependencies and remain invariant over the admitted input domain; (H3) dependencies, storage ownership and materialisation ensure that each consumer reads the intended value; and (H4) the numerical configuration remains fixed. Under these conditions, equal input and constant bits give equal predecessor values. Corresponding operations produce equal result bits, and induction over the dependency graph reaches the outputs.
Hoisting is valid only for computations that are independent of changing
inputs and mutable state, and the resulting bits must be retained. Reordering
must preserve data and effect dependencies, including asynchronous completion.
Storage reuse must preserve ownership, alias assumptions and every consumer's
lifetime. A high-level convolution with workspace-dependent implementation
selection is not itself a fixed bit-semantic operation, and launch capture
does not remove these obligations. This reasoning follows established
specialisation and translation-validation principles
\citep{Jones1993PartialEvaluation,necula2000translation,lopes2021alive2}.
The implementation does not mechanically verify every condition.

\subsubsection*{Deterministic components and retained-reference agreement}
A deterministic replacement and a reference-preserving replacement satisfy different requirements. TensorRT documentation distinguishes built-engine repeatability from build-time implementation choices, subject to runtime conditions and exceptions \citep{nvidia2026tensorrtdeterminism}; repeatability does not by itself establish agreement with a retained eager reference. CUB single-phase reduction APIs provide selectable run-to-run and GPU-to-GPU determinism for supported reductions \citep{nvidia2026cccldeterminism}, but do not define historical equality for a complete inference graph. We use these mechanisms to distinguish component determinism from complete-output agreement; they were not timed in the paced-service cohort.

The empirical pass is a diagnostic comparator, not the primary reuse mechanism. It intercepts selected CUDA tensor constructors, host-to-device
transfers and host-provided indexing. It keys observations by source site and
occurrence order, requiring a matching nonzero occurrence count and equal
values across qualification inputs. Scalar extraction is disabled. The pass
does not discover arbitrary invariant arithmetic subgraphs, which is why the
dense trunk layers require explicit specialisation.

The inventories agree across all 27 comparator blocks. DeepONet empirical capture substitutes two uniform-coordinate constructors and one $2\pi$ constructor; explicit trunk reuse uses no empirical sites. FNO substitutes one spectral-buffer constructor site with four occurrences. The FNO constructor initially returns zeros, but its storage is subsequently modified. Reusing this scratch buffer therefore requires an overwrite and initialisation argument rather than an immutable-constant claim. Source inspection identifies the relevant dependencies and writes; finite comparisons alone do not establish alias safety or all-input dependency validity. Trunk tables and empirical cache copies are created under inference mode. Their signatures record identity and metadata, but a missing version counter is recorded rather than rejected. Contents are not compared by the runtime checks, so untracked in-place writes remain outside those checks \citep{pytorch2026inferencemode}. H2 concerns reused values; H3 additionally requires correct initialisation and every read to observe the intended write when storage itself is reused.

\subsubsection*{Explicit execution controls and mutable-storage audit}
The paced-service comparison removes empirical constancy inference from its primary
efficiency paths (Table~\ref{tab:si-control-costs}). Literal DeepONet capture
failed on an unpinned host-to-device scalar transfer. Explicit preparation
of the fixed $2\pi$ tensor enabled capture; it did not substitute the grid
arithmetic or infer a constant from witnessed payloads. Native FNO capture
succeeded with the native scratch construction and initialisation.

The existing DeepONet reuse path and the empirical-disabled path each returned 32 exact outputs in a diagnostic. Their 22 ordered GPU event names and recorded allocated/reserved storage agreed. Only $G_T$ enters the paced-service performance matrix. The diagnostic compares the recorded traces and storage, without establishing equal preparation costs, launch parameters or behaviour on other inputs.

For the FNO empirical comparator, a separate actual GPU probe identified
four cached occurrences at the spectral-constructor source site. Each had
shape $[1,40,94,48]$, complex64 dtype, 1,443,840 bytes and distinct contiguous
storage. The native 16-mode bands occupied rows 0--15 and 78--93 and
columns 0--15, so the two written bands were disjoint. Across eight
snapshots, each occurrence retained 160,000 complex elements of bitwise
positive zero outside those bands. All six A--B--A--B--A--B outputs matched
their saved targets. The enumerated cache buffers, exposed input/output
storage and model state did not overlap.

The cached contents changed between A and B, while inference-mode version counters were unavailable. These buffers are mutable scratch storage rather than immutable reused values. Source-defined assignment order and modelled overwrite/alias controls support the effect analysis. Private captured temporaries, occurrence-to-layer mappings and protection against injected cache corruption were not enumerated by this diagnostic. Native $G_0$ provides the measured alternative without the scratch substitution.

In the paced-service implementation, state and resource checks run after owned-input decoding and again after synchronisation and complete output materialisation. They inspect numerical settings, module state, parameter and buffer identities, available version counters, public configuration and owned-storage metadata. Observed allocation-retry or out-of-memory history prevents preparation, qualification or service; these events do not themselves diagnose numerical disagreement \citep{pytorch2026memorystats}. An error or timeout closes the serving path. Recovery retires the worker, evaluates the specified resource-readiness condition and qualifies a replacement against the original target; failed requests are not replayed automatically. The measured replacement path uses $E_{\rm bit}$, while relaxed relations were evaluated at startup only.

The eager, graph and eligible reuse arms share qualification, guards, transport, complete CPU output and the offered request trace. Their contrasts measure the execution strategy under this common policy. Separate controls hold execution and output ownership fixed while adding supervision (Supplementary Note~\ref{si:legacy}); their costs apply to that prepared-device interface.

Determinism constrains repeatability \citep{pytorch2026randomness,heumos2023mlfcore,horvat2022wiff}, while numerical regression compares an execution with saved outputs or reference ensembles \citep{nvidia2026polygraphy,baker2015cect,e3sm2026eamxxtesting}. Numerical contracts specify kernel requirements \citep{veit2026kernelcontracts}; equivalence verification bounds network-output differences \citep{mo2026eqbab}, and translation validation checks source-to-target refinement under a specified semantics \citep{pnueli1998translation,lopes2021alive2}. InfraValidator tests model loading and serving before deployment \citep{tensorflow2024infravalidator}; Kubernetes probes gate traffic and restart unhealthy containers \citep{kubernetes2026probes}. Numerical checks remain application-specific. OpCert retains the target across these operations while using established graph and fixed-dependency reuse mechanisms \citep{pytorch2021cudagraphs,Jones1993PartialEvaluation}.

\OpCertSInote{si:history}{Resource history and worker recovery}
\label{si:reference-recovery}

These controlled interventions test execution-state dependence, rather than an energy-saving policy or a claim that services should vary allocator caps. The paced-service resource-history comparison holds the checkpoint, retained reference
and cyclic serving trace fixed while changing memory ownership. Workers
are admitted at 1,536~MiB. Each completes six high-cap requests, six requests
at 320~MiB without explicit reclaim, and six scheduled requests at the same
cap after unused-pool release, followed by three restored or replacement
requests. The six-request phases alternate the same two serving positions.
The phases are an ordered history, not independent factorial samples.

Ordinary eager and the existing prepared graph have three fresh-worker
episodes each. Dedicated-pool native eager is an additional three-episode
control. It retains a reusable \texttt{MemPool} context from initialisation
through close, without graph replay or occupied dummy tensors. The same
\texttt{empty\_cache} call acts on different reclaim-eligible storage under
these policies \citep{pytorch2026cudasemantics}. The control compares ownership
policies, not equal bytes physically released.

\begin{table}[htbp]
\centering\small
\caption{\textbf{Storage-policy control on RTX~5060~Ti.} Counts sum three worker episodes per policy with the same ordered input sequence. Refusal ends that phase. Dedicated-pool episodes are a subsequent control, not paired performance blocks.}
\label{tab:si-storage-control}
\setlength{\tabcolsep}{4pt}
\begin{tabular}{@{}p{0.26\linewidth}rrrr@{}}
\toprule
 & High cap & Low, retained & Low, reclaim & Restore / recover \\
\midrule
Ordinary eager & 18 exact & 18 exact & 3 refusals & 9 exact \\
Dedicated-pool eager & 18 exact & 18 exact & 18 exact & 9 exact \\
Prepared graph & 18 exact & 18 exact & 18 exact & 9 exact \\
\bottomrule
\end{tabular}
\end{table}

All 171 returned fields matched their retained targets. Dedicated-pool eager kept 370~MiB reserved and 75.624~MiB allocated across the recorded boundaries, with 13 segments and no increase in cumulative device-allocation counts. No restart was required. Graph replay was therefore unnecessary for continuation under this history. The result does not imply a 320~MiB physical working set or zero tensor allocation.

Ordinary eager refused on recorded allocation-failure history after reclaim. A high-budget replacement requalified against the original target and returned an exact output in 1.814, 1.849 and 1.806~s. The timer includes worker retirement and ends at delivery, before numerical comparison or a diagnostic memory query. The readiness predicate records restoration of the synthetic admission budget rather than external free memory. Because refused requests produced no candidate arrays, their numerical relation to the reference is unknown.

The completed Jetson eager episode returned 15 exact fields and recovered in 5.610~s. Graph admission failed for allocation-retry history at 1,536~MiB; the stop rule left later groups unattempted. This is an incomplete extension, not a three-group comparison or a minimum-memory estimate. A preceding environment-setup failure is excluded from these outcomes and retained in the execution archive.

\subsubsection*{Resource-history service at a different cap}
The resource-history service cohort used the released shear-flow UNetClassic checkpoint from The Well \citep{ohana2024well}. It is distinct from the Rayleigh--B\'enard plan-selection example below. Two of its four retained inputs qualify the worker and two are served. Three paired RTX~5060~Ti eager/graph groups tested a 304~MiB allocator cap applied after admission at 1,536~MiB. All 42 returned fields matched the original reference; three eager resource refusals remain separate availability outcomes. Eager recovery takes 1.781, 1.789 and 1.789\,s; graph returns all six restricted requests per group without replacement. The Jetson eager episode returns four exact fields and recovers in 5.942\,s. Jetson graph admission fails for allocation-retry history at 1,536\,MiB, leaving later groups unattempted. Recovery times end at the first valid replacement reply.

The local cap does not evict allocations retained during preparation. Continued graph execution therefore does not establish a 304~MiB physical working set or immunity to external memory pressure. The eager refusals record allocation-failure history rather than numerical corruption. Because no candidate arrays were returned, no numerical mismatch is assigned to those refusals. Recovery retains the original reference and requires qualification of the replacement worker. Releasing allocator memory alone does not reset the execution-plan cache.

The Jetson graph result is an admission limit of the tested configuration. Earlier launcher failures are excluded; neither attempt produced successful low-cap graph service.

A separate input-history control tested stale intermediates without changing workers or applying memory pressure. Each eager, graph and reuse path served $A\rightarrow B\rightarrow A\rightarrow C\rightarrow A$, where $A$ is an existing serving input, $B$ reflects one coordinate-valued branch feature and $C$ interpolates the coefficient channel. The comparator used previously saved, twice-repeatable eager outputs. All 15 requests matched their targets, including all six later returns to $A$, and previously returned replies remained unchanged after subsequent requests. The probes test input association and storage reuse on these five requests.

The coordinate-valued payload channels and the trunk query grid are distinct.
For the evaluated GridDeepONet2d, the branch interpolates and flattens all
three payload channels, $z=(a,\xi,\eta)$. The native trunk grid $X_{h,w}$ is
generated independently from the admitted shape, device and dtype by
uniform coordinate construction. Fixed-trunk specialisation retains the
same native grid and original trunk chunks. Probe B replaces payload channel
1 by one minus its value; probe C interpolates channel 0 between two existing
coefficient inputs. Neither changes $X_{h,w}$. These probes therefore test
branch-input association and stale intermediates, not support for arbitrary
new trunk queries or physical geometries.

The retained arrays also establish that B and C require different answers from A. B differs in 177,231 of 177,241 normalised output elements, with maximum absolute difference 0.00632119 and relative $L_2$ difference 0.00203527. C differs in all 177,241 elements, with corresponding differences 0.65312338 and 0.23316280. The three arms reproduce both modified-input targets and the original answer on every return to A. The B/C targets were generated for this normalised-output comparison; physical truth is unavailable for the modified inputs.

\subsubsection*{Resource intervention and recovery policy}
Supplementary Fig.~\ref{sifig:pressurecontrols} shows the allocator-cap sweep
and places the execution difference beside the physical prediction error;
Supplementary Table~\ref{si:tab-controls} separates numerical-setting and
external-occupancy controls. Supplementary
Fig.~\ref{sifig:memoryhistory} shows the 18 shear-flow histories: three storage/cache policies, each repeated in three fresh processes on RTX~5060~Ti and GH200. Nine additional seeded Rayleigh--B\'enard histories bring the complete control set to 27 processes. Discrepancies use each process's initial output. Complete allocator snapshots accompany the archived trajectories. These controls are separate from the paced-service cost comparison.

\setcounter{figure}{0}
\begin{figure}[!htbp]
\centering
\includegraphics[width=\linewidth]{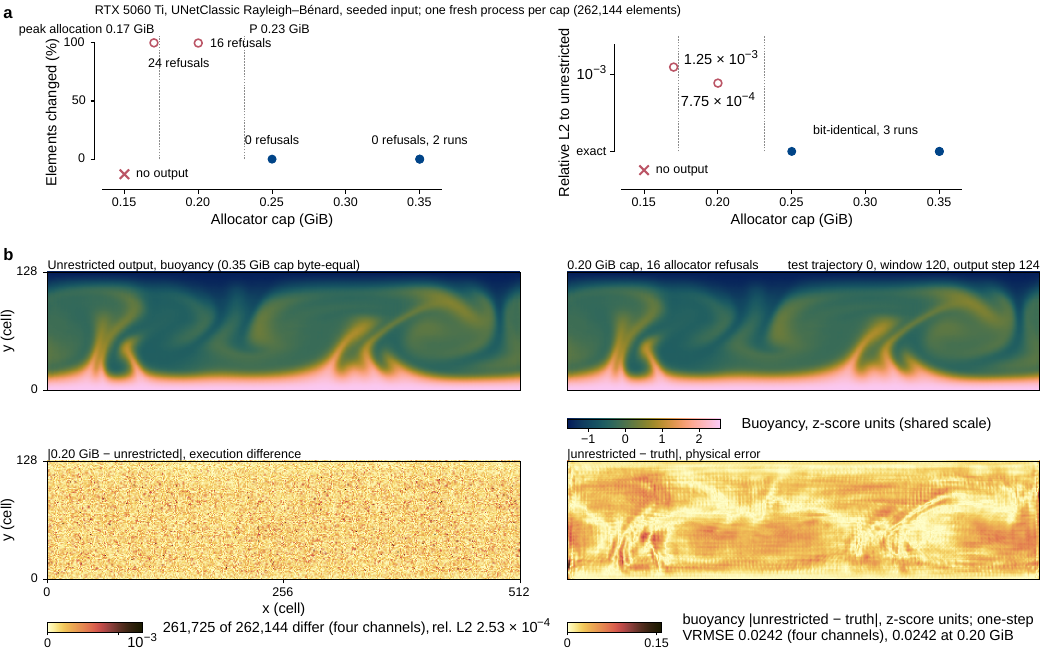}
\caption{\textbf{Allocator-cap sensitivity and physical prediction error.}
\textbf{a}, Rayleigh--B\'enard on RTX~5060~Ti, one fresh process per cap: fraction of changed elements and relative $L_2$ to the unrestricted output. Crosses mark no output at 0.15\,GiB. The pressure scale is $P=A_{\mathrm{peak}}+W_{\max}$.
\textbf{b}, Test window 120: unrestricted and capped predictions, their absolute difference, and unrestricted prediction error against the next state. Both predictions have one-step VRMSE 0.0242. Numerical-setting and external-occupancy controls are in Supplementary Table~\ref{si:tab-controls}.}
\label{sifig:pressurecontrols}
\end{figure}

\setcounter{figure}{1}
\begin{figure}[!htbp]
\centering
\includegraphics[width=\linewidth]{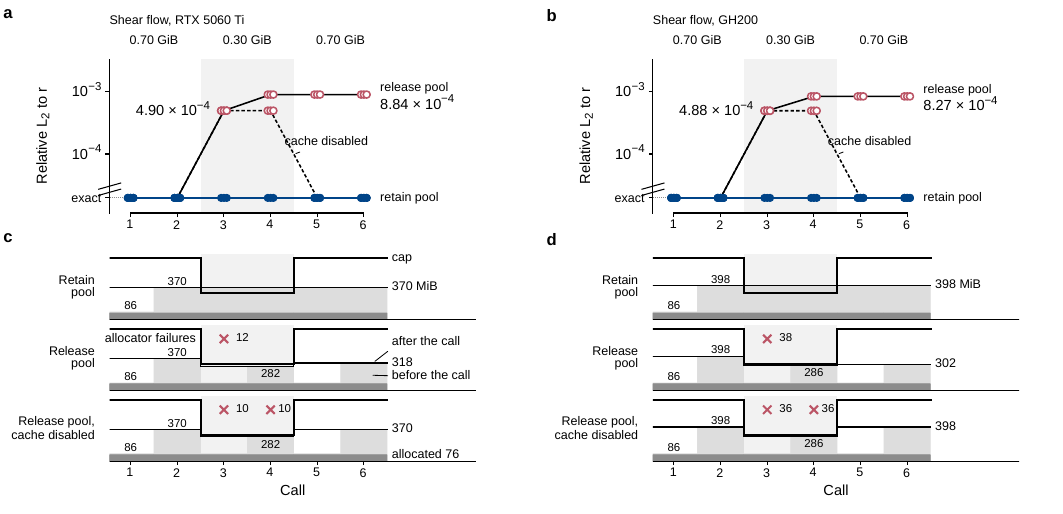}
\caption{\textbf{Storage retention and plan-cache policy separate continuation from recovery.}
\textbf{a,b}, Relative $L_2$ to the first output over six calls on RTX~5060~Ti and GH200. Caps change from 0.70 to 0.30 to 0.70\,GiB. Three processes per policy coincide. After restoration, pool-release outputs remain different with the cache enabled but recover the initial output with the cache disabled.
\textbf{c,d}, Reserved and allocated memory per call. Labels give low-cap allocator refusals (RTX: 12 and 10; GH200: 38 and 36). There are nine fresh processes per device.}
\label{sifig:memoryhistory}
\end{figure}

\input{si/resource_history_detail}

\begin{table}[!htbp]
\centering\small
\caption{\textbf{Numerical-setting and external-occupancy controls.} Seeded Rayleigh--B\'enard, fixed checkpoint and GPU. Rows 1--3 compare a 0.20\,GiB cap with unrestricted execution; row 4 uses a 0.35\,GiB reference. The final row applies external occupancy without an allocator cap. Counts describe this single-input diagnostic.}
\label{si:tab-controls}
\InputIfFileExists{si_table_numerical_controls}{}{\input{si/si_table_numerical_controls}}
\end{table}

\OpCertSInote{si:contractfrontier}{Alternative numerical criteria and recovery controls}

These controls examine alternative output criteria; they do not relax the byte-identity constraint of the main energy comparisons. The Darcy transition experiments hold each checkpoint, numerical configuration and 24-input bank fixed. Positions 0--7 supply qualification witnesses and positions 8--23 supply evaluation inputs. Fresh processes evaluate each input twice; references must be finite and byte-repeatable. Candidate outputs, construction failures, hashes and versions are retained. The subsequent service comparison uses resident input-bank indices and complete CPU replies, separately from the 72-block full-input service in Note~\ref{si:service}.

\subsubsection*{Numerical predicates and their interpretation}
The byte predicate $E_{\rm bit}$ is Eq.~\ref{eq:si-output-contract}. For finite normalised candidate output $c$ and retained output $r$, $E_{\rm num}$ requires $|c-r|\leq10^{-6}+10^{-5}|r|$ for every element. These predicates concern the realised execution, independently of prediction error against a physical target.

The field predicate uses the unchanged CPU decoder $D$ and native-grid simulated truth $y^*$:
\begin{equation}
 \rho_{\rm exec}=\frac{\|D(c)-D(r)\|_2}{\|D(r)-y^*\|_2},\qquad
 \rho_{\nabla}=\frac{\|\nabla_hD(c)-\nabla_hD(r)\|_2}
 {\|\nabla_hD(r)-\nabla_hy^*\|_2}.
 \label{eq:si-field-predicate}
\end{equation}
Here $\nabla_h$ concatenates centred interior finite differences on the native unit-square grid. Norms use float64 after unchanged float32 decoding. $E_{\rm field}$ requires both ratios to be at most $\eta=0.01$ for every witness. A zero numerator and denominator give ratio zero; a positive numerator with zero denominator, or any nonfinite comparison, is rejected. This gradient diagnostic is not a PDE-residual test.

For pressure $p$ and permeability $K$, the section flux is
\begin{equation}
 Q(p,K)=-\sum_{x\in\mathcal X_h^{\rm int}}
 K(y_0,x)(\nabla_h p)_y(y_0,x)\,\Delta x,
 \qquad y_0=0.5,
 \label{eq:si-flux-predicate}
\end{equation}
where $\Delta x$ is the native-grid spacing, $\mathcal X_h^{\rm int}$ contains the interior section-grid locations and permeability uses the checkpoint's input-grid indices. The flux predicate is $E_{\rm flux}: |Q_c-Q_r|\leq0.01|Q_r-Q_{\rm truth}|$, with $Q_c$, $Q_r$ and $Q_{\rm truth}$ evaluated on the decoded candidate, reference and simulated truth. It constrains that section, not full conservation. Integration can cancel local differences, whereas differentiation can amplify small-scale changes, so field, gradient and flux criteria are not interchangeable.

For a field, gradient or scalar flux $v$, let $v_c$, $v_r$ and $v^*$ denote the candidate, reference and simulated truth under the same interpretation. The triangle inequality gives, for $\eta\geq0$,
\begin{equation}
 \|v_c-v_r\|\leq\eta\|v_r-v^*\|
 \ \Longrightarrow\ 
 \|v_c-v^*\|\leq(1+\eta)\|v_r-v^*\|.
 \label{eq:si_error_budget}
\end{equation}
At $\eta=0.01$, the candidate error is bounded by 1.01 times the reference error for the same truth, grid and interpretation. Zero reference error requires zero discrepancy; a less accurate reference permits larger absolute changes. These thresholds were fixed before candidate comparison as illustrative experimental budgets, not application-derived acceptance requirements. Rejected candidates retain offline results but do not yield admitted replies.

\subsubsection*{Candidate outcomes and live admission}
\begin{table}[htbp]
\centering\small
\caption{\textbf{Qualification and evaluation outcomes for runtime transitions.} Each entry gives passing witnesses out of eight, followed by passing evaluation inputs out of 16. Candidates are compiled PyTorch except the specialised Jetson TensorRT engine. Counts are offline output comparisons; live gates are reported in Supplementary Fig.~\ref{sifig:relations}c.}
\label{tab:transition-counts}
\begin{tabular}{@{}lrrr@{}}\toprule
Device and candidate & $E_{\rm bit}$ & $E_{\rm num}$ & $E_{\rm field}$ \\\midrule
5060 Ti, DeepONet & 0/8; 0/16 & 8/8; 16/16 & 8/8; 16/16 \\
5060 Ti, FNO & 0/8; 0/16 & 0/8; 0/16 & 8/8; 16/16 \\
A2000, DeepONet & 0/8; 0/16 & 8/8; 16/16 & 8/8; 16/16 \\
A2000, FNO & 0/8; 0/16 & 0/8; 0/16 & 7/8; 15/16 \\
Jetson, TensorRT & 0/8; 0/16 & 8/8; 16/16 & 8/8; 16/16 \\
\bottomrule\end{tabular}\end{table}

Both compiled models differed from eager on every evaluation input on RTX~5060~Ti and A2000 (Table~\ref{tab:transition-counts}). The A2000 FNO candidate failed one qualification and one evaluation input under $E_{\rm field}$. Its maximum evaluation gradient ratio was 0.0101013, just above the prescribed 0.01 boundary, while its maximum field ratio was 0.00315018. Native FNO graph replay and DeepONet fixed-trunk execution preserved all evaluated bytes on the three devices. Direct DeepONet capture failed; capture after empirical hoisting succeeded.

\emph{Live admission.} Startup qualification evaluates each of eight witnesses twice under the declared relation before the worker reaches READY. Any failed comparison refuses admission. On RTX~5060~Ti, $E_{\rm bit}$ refuses both compiled models; $E_{\rm num}$ admits DeepONet only; $E_{\rm field}$ and $E_{\rm flux}$ admit both. Evaluation-position comparisons audit an admitted service and do not substitute for this startup gate.

\emph{Physics-derived flux relation.} The middle section and original 1\% predicate (Eq.~\ref{eq:si-flux-predicate}) were fixed before their candidate comparisons. Both models passed on all 16 served positions; maximum $\rho_{\rm flux}=|Q_c-Q_r|/|Q_r-Q_{\rm truth}|$ was $3.86\times10^{-5}$ for DeepONet and $7.71\times10^{-3}$ for FNO. The latter uses 77\% of its budget, compared with 2.6\% and 8.3\% for field and gradient changes ($2.56\times10^{-4}$ and $8.35\times10^{-4}$). The three ratios use different reference-error denominators. A subsequent check fixed two further sections at 0.45 and 0.55 before evaluating them on the saved outputs. For requested fraction $y$, the code uses zero-based row $\operatorname{round}((n-1)y)$, clipped to the interior rows $[1,n-2]$. Fractions 0.45, 0.50 and 0.55 give rows 189, 210 and 231 on the $421^2$ grid and 38, 42 and 46 on the $85^2$ grid. Flux is evaluated on these native rows without interpolation. All three sections passed for both models (16/16 each); section-wise maximum ratios ranged from $3.1$--$5.7\times10^{-5}$ for DeepONet and $1.9$--$7.7\times10^{-3}$ for FNO. This section check used evaluation positions only and did not rerun startup qualification.

\begin{figure}[!htbp]
\centering
\includegraphics[width=\linewidth]{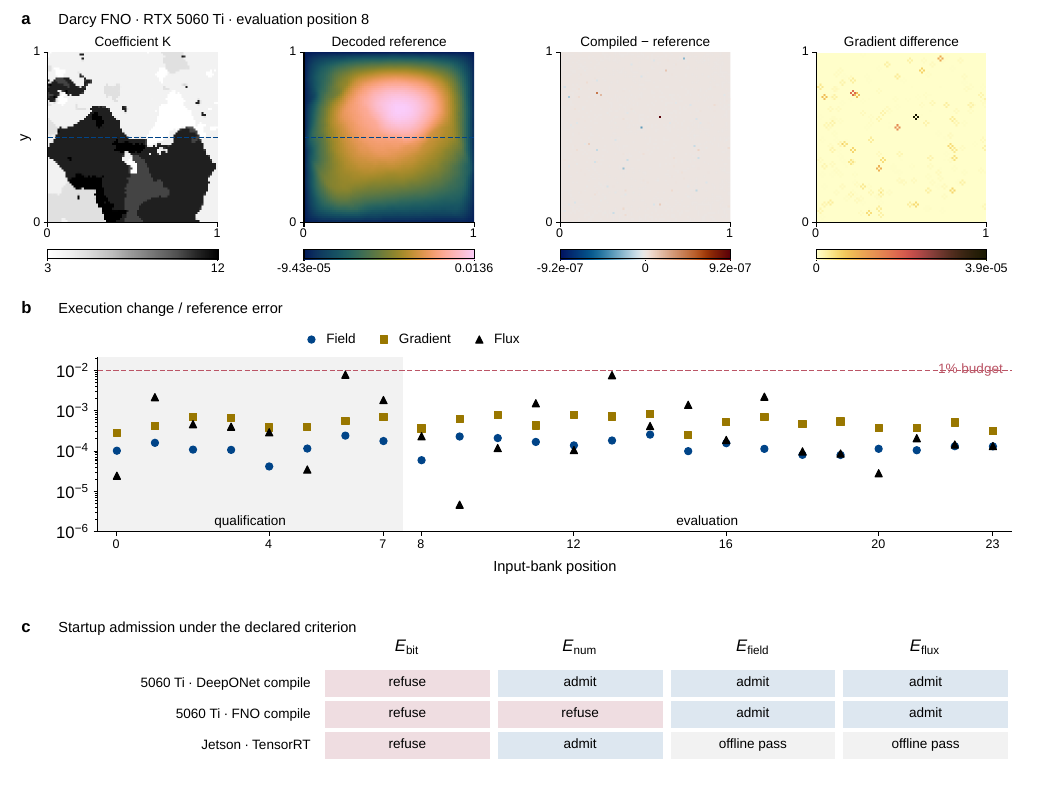}
\caption{\textbf{Illustrative field and flux budgets change admission decisions.}
\textbf{a}, Darcy FNO on RTX~5060~Ti, evaluation position 8: coefficient, decoded reference, compiled--reference field difference and gradient-difference magnitude. The line marks the flux section at $y=0.5$.
\textbf{b}, FNO execution changes relative to reference errors on RTX~5060~Ti for eight qualification and 16 evaluation inputs. Colours and markers distinguish field, gradient and flux; the dashed line is $\eta=0.01$.
\textbf{c}, Live startup decisions; grey cells denote offline evaluation. Offline counts for the bitwise, numerical and field criteria, including A2000 FNO failures, are in Table~\ref{tab:transition-counts}. These budgets are not application acceptance thresholds.}
\label{sifig:relations}
\end{figure}

\begin{table}[!htbp]
\centering\small
\caption{\textbf{Contemporaneous service measurements of qualified candidates.}
Six fresh-worker blocks per arm on RTX~5060~Ti, alternating AB/BA order within each model's session. Entries are arm medians with observed min--max ranges. The relation column gives the gate used in the measured blocks; separate admission tests are reported in Supplementary Fig.~\ref{sifig:relations}c.}
\label{si:tab-frontier}
\begin{tabular}{@{}lllrr@{}}
\toprule
Model & Candidate & Gate & Response (ms) & Energy (J/request) \\
\midrule
DeepONet & Fixed-trunk graph & $E_{\rm bit}$ & 1.43 (1.42--1.45) & 0.339 (0.335--0.343) \\
DeepONet & Compile & $E_{\rm num}$ & 9.18 (9.13--9.22) & 1.604 (1.593--1.613) \\
FNO & Native graph & $E_{\rm bit}$ & 0.780 (0.774--0.790) & 0.0882 (0.0876--0.0886) \\
FNO & Compile & $E_{\rm field}$ & 1.28 (1.25--1.29) & 0.0930 (0.0928--0.0934) \\
\bottomrule\end{tabular}
\end{table}

Both arms were measured in the same session, using six paired groups and 12\,s arrival horizons. DeepONet and FNO received 864 and 8,720 requests per block, respectively. Qualification and preparation preceded timing. The worker receives an input index and reads the corresponding tensor from its resident CPU bank. The exact arm checks every output before publication, whereas compile and TensorRT omit that check. The client records byte agreement inside the timed loop and retains arrays for subsequent field/flux analysis. The service measurements include these unequal guard costs. Individual blocks are retained in the source data. Earlier unmatched measurements are archived but are not pooled into Table~\ref{si:tab-frontier}. Flux qualification was tested separately; changing the startup predicate did not define a further timed execution path. The latency and energy ranges do not overlap within the measured service interval; preparation and recovery costs are reported separately.

\emph{Recovery ablation.} Each Darcy model was evaluated under a weight-mutation fault in three configurations (Table~\ref{si:tab-lifecycle}). The retained-target pilot used a per-request guard and explicitly invoked worker replacement and requalification after refusal. The ablation used the same guard but withheld recovery; all three post-fault attempts were refused, the first on the mutation and the next two because the service remained quarantined. The weight-generation guard detects the mutation before execution. A separate periodic-check control served 16 requests after the mutation before comparing the collected outputs with the retained answers. All 16 differed. Each configuration was run once per model on RTX~5060~Ti; these are mechanism controls without latency or energy attribution.

\begin{table}[!htbp]
\centering\small
\caption{\textbf{Service state after weight mutation with recovery invoked or withheld.}
RTX~5060~Ti, one weight-mutation run per model and configuration. Both models gave the same qualitative outcome; the retained-target row uses the archived guarded pilot.}
\label{si:tab-lifecycle}
\begin{tabular}{@{}llrl@{}}
\toprule
Configuration & Detection & Incorrect replies & Recovery outcome \\
\midrule
Guard with recovery & Before publication & 0 & Ready after requalification \\
Recovery withheld & Before publication & 0 & QUARANTINED \\
Periodic check & After window closes & 16 & Not attempted \\
\bottomrule\end{tabular}
\end{table}

Recovery was exercised under $E_{\rm bit}$ only. Replacement without requalification was not evaluated, so the ablation tests the combined procedure. The periodic window was fixed before the run. A separate comparison with an independently configured recovery service is reported in Supplementary Note~\ref{si:edge-service}.

\emph{Cross-runtime, cross-device extension.} Full-network Darcy DeepONet TensorRT construction failed, and the tested FNO ONNX export failed at its FFT operator. The initial Jetson eager-preparation failure is retained separately in the availability record. The successful fixed-shape DeepONet engine used retained trunk tables and was repeatable on all 24 inputs. The fixed-trunk tables were generated under the Jetson PyTorch configuration, copied without recomputation and hashed into a fixed-shape ONNX opset-17 export. TensorRT~10.7 was built with \texttt{trtexec}; engine execution bound device buffers through \texttt{set\_tensor\_address} and \texttt{execute\_async\_v3}, synchronised and retained CPU outputs. Earlier in-process optimisation-level-3 full-network and specialised builds failed. Process isolation and optimisation level changed together for the successful build, so their effects are not separately identified.

Jetson reused the qualified TensorRT fixed-trunk DeepONet engine unmodified (builder optimisation level 0, FP16/TF32 disabled, 256\,MiB workspace). Its live gate refused $E_{\rm bit}$ and admitted $E_{\rm num}$. Field and flux checks passed offline on all 16 served positions; their decoder and truth were available on the analysis host, and no corresponding live gates were implemented on Jetson. The contemporaneous comparison used three fresh-worker blocks per arm, two AB groups and one BA group. TensorRT gave median response 6.79\,ms (6.59--6.87) and request energy 0.888\,J (0.886--0.889), against 5.42\,ms (5.21--5.59) and 0.866\,J (0.866--0.869) for the byte-qualified PyTorch fixed-trunk graph. These ranges retain the different guard costs of the two configured services.

\emph{Qualification-record exports.} A versioned qualification-record schema consolidates the checkpoint, interface, configuration, retained corpus, numerical predicate, evidence, monitored conditions and recovery policy. CPU-only utilities export an $E_{\rm num}$ compiled-service record and an $E_{\rm bit}$ recovery-pilot record. Validation checks required fields, artifact hashes, reference and relation identities, witnessed positions and consistency with archived admission and recovery events; it does not rerun inference. The original services did not consume these exported manifests, and the compiled-service example has recovery disabled.

\input{si/temporal_scope}

%% file: si/cohort_map.tex
\ifdefined\OpCertSIStandalone\clearpage\fi
\nolinenumbers
\begingroup\setstretch{1}\small
\noindent\textbf{Guide to the Supplementary Information}\par\vspace{.7em}
The comparisons below use different input/output and energy boundaries. Embedded-device energy measures module input; the desktop measurements in Fig.~\OpCertMainRef{fig:service} measure GPU energy. All include waiting within the stated interval, without idle subtraction.\par\vspace{.7em}
\setlength{\tabcolsep}{4pt}
\begin{tabular}{@{}p{.17\linewidth}p{.27\linewidth}p{.50\linewidth}@{}}\toprule
Comparison & Input/output boundary & Energy interval and numerical comparison \\\midrule
Fig.~\OpCertMainRef{fig:source_edge}\newline Note~\ref{si:source-edge} & Raw CPU input to decoded CPU fields & 20\,s continuous inference at the retained device configuration. Preparation and 10\,s warmup precede timing; startup checks precede timing and complete output banks are compared after each block. \\
Fig.~\OpCertMainRef{fig:edge_sustained}\newline Note~\ref{si:edge-service} & Raw CPU input to normalised and decoded CPU fields & Preparation through closure, including 20\,s warmup and 120\,s of arrivals. Post-delivery comparisons are timed. Power/clock policies are compared separately; shaded conditions complete unequal work. \\
Fig.~\OpCertMainRef{fig:service}\newline Note~\ref{si:service} & Normalised CPU input to normalised CPU fields & Input encoding through delivery, spanning at least the 12\,s arrival horizon. Preparation, warmup of at least 5\,s, physical decoding and post-block output comparison are excluded. \\
Fig.~\OpCertMainRef{fig:operating_cost}\newline Note~\ref{si:edge-service} & Same heat-exchanger service interface as Fig.~\OpCertMainRef{fig:edge_sustained} & Preparation through closure at 15\,W automatic clocks. Qualification and capture are included; no additional warmup is added. Post-delivery comparisons are timed. Artifact construction is measured separately. \\
Fuel-cell replay\newline Note~\ref{si:temporal-scope} & Recorded observations to normalised and decoded CPU fields & Common 120.998\,s window including waiting, checks and replacement. Initial preparation and final shutdown are excluded. Returned fields are saved for later comparison and held-field assessment. \\
\bottomrule\end{tabular}
\vspace{.8em}\par
\noindent\textbf{Supporting controls.} Notes~\ref{si:profiling}--\ref{si:legacy} give profiling, physical-output audits and the separate configuration/device studies. Notes~\ref{si:scope}--\ref{si:history} define reference scope, preservation conditions and allocator-history controls. Note~\ref{si:contractfrontier} evaluates alternative numerical criteria. The serving-framework comparison is in Note~\ref{si:edge-service}; the distinction between computational reuse and holding an earlier prediction is in Note~\ref{si:temporal-scope}.\par\vspace{.6em}
\noindent\textbf{Units of evidence.} Repeats use fresh processes or workers on the same device; requests and field locations are not additional replicates. Paired summaries report medians and observed ranges, not confidence intervals. Energy contrasts use the same completed request sequence unless unequal completion is stated. Repeated power measurements and integration checks do not establish sensor calibration uncertainty.\par\vspace{.6em}
Full records retain failed and unavailable candidates, individual blocks and the 127-case allocator matrix. The Well checkpoint supports execution-history controls; its recovered physical predictions underperform persistence (Note~\ref{si:physical}).
\endgroup
\clearpage


%% file: si/source_to_edge.tex
\OpCertSInote{si:source-edge}{Heat-exchanger execution costs and source comparison}

This note supports Fig.~\OpCertMainRef{fig:source_edge}: it separates agreement among execution paths on Jetson from agreement with the workstation reference, then reports inference costs. The released deterministic MIMONet heat-exchanger model and the original 310 test cases define the comparison.

\subsubsection*{Reference and physical-output comparison} The source reference is newly designated on RTX~5060~Ti after reproducing the released batch-eight accuracy evaluator. The batch-one reference was repeated twice. The checkpoint was obtained from the archived MIMONet release \citep{kobayashi2026mimonetdata}; local data payload sizes and CRC32 values match its retained ZIP central directory. File SHA256 values, exact code versions and extraction provenance are retained. The complete 3.74\,GB archive checksum was not independently verified. The checkpoint, split, decoder and geometry were fixed before edge evaluation.

The evaluator selects raw target columns $[1,3,4,5]$, ordered as pressure, $u_z$, $u_y$ and $u_x$. Its branches receive the scalar inlet pair and the 100-point heat-flux profile. The coordinates are scaled independently to $[-1,1]$; input normalisation and the original CPU inverse scaler are retained.

All four Jetson arms failed source byte identity on eight startup witnesses and on all 302 subsequent evaluation cases. The 16 saved output banks were mutually byte-identical, before and after physical decoding. Every bank was independently reanalysed against the source and CFD target. Supplementary Table~\ref{tab:source-edge-errors} reports the resulting error changes. The published accuracy table and the official reproduction tolerance of 0.05 percentage points were used only to check recovery of the original model evaluation; neither was adopted as a deployment acceptance rule.

\begin{table}[!htbp]\centering\small
\caption{\textbf{Physical-output comparison on all 310 original test cases.} Relative $L_2$ entries are means in percent. Maximum field differences use Pa for pressure and m\,s$^{-1}$ for velocity. The final column is the largest per-case increase in relative error, in percentage points. All Jetson arms share these arrays.}
\label{tab:source-edge-errors}
\begin{tabular}{@{}lrrrr@{}}\toprule
Field & Source error (\%) & Jetson error (\%) & Max. difference & Max. error increase \\\midrule
$p$ & 0.8188057 & 0.8188016 & 0.017578 & 9.16e-06 \\
$u_z$ & 1.4538439 & 1.4538449 & 7.6294e-06 & 7.7e-06 \\
$u_y$ & 1.0175234 & 1.0175252 & 1.1273e-05 & 3.98e-06 \\
$u_x$ & 0.5205477 & 0.5205475 & 2.6464e-05 & 1.44e-06 \\
\bottomrule\end{tabular}\end{table}

The source uses PyTorch~2.13.0+cu129 and NumPy~2.2.5; the edge uses PyTorch~2.5.0a0+872d972e41.nv24.08 and NumPy~1.23.5. The decoder is part of the compared execution, with the same inverse-scaler source and constants in both environments. Applying these two CPU stacks to the same saved normalised source array gives a maximum absolute difference of 0.00390625 in the decoded tensor. The actual GPU-normalised predictions also differ, by at most $5.72205\times10^{-6}$ across all entries. Both GPU inference and CPU decoding therefore contribute possible numerical differences across the two software stacks; this diagnostic does not separate their contributions additively.

\subsubsection*{Continuous-inference costs and device configuration}

\begin{table}[!htbp]\centering\small
\caption{\textbf{Continuous-inference costs at the recorded Jetson configuration.} Medians and observed ranges over four process blocks per path; latency is the median over requests within each block. E, eager; G, graph; R, trunk reuse; RG, reuse with graph. Memory is peak CUDA allocator usage, not total module RAM. All paths preserve the evaluated local outputs.}
\label{tab:source-edge-costs}
\begin{tabular}{@{}lrrrr@{}}\toprule
Arm & Latency (ms) & Energy (J/request) & Allocated (MiB) & Reserved (MiB) \\\midrule
E & 6.527 [6.492, 6.567] & 0.1087 [0.1062, 0.1089] & 70.35 & 82 \\
G & 4.439 [4.433, 4.452] & 0.0886 [0.0872, 0.0888] & 70.82 & 112 \\
R & 3.360 [3.294, 3.370] & 0.0430 [0.0427, 0.0434] & 70.35 & 82 \\
RG & 1.453 [1.448, 1.467] & 0.0258 [0.0254, 0.0259] & 118.82 & 140 \\
\bottomrule\end{tabular}\end{table}

Reuse retains 16,289,792 bytes (15.5352\,MiB) of local trunk features. Counting a multiply and add as two operations, the four removed dense layers account for approximately 3.132\,GFLOP per request, computed from the layer shapes. Source features are not substituted for the locally computed table. Construction and startup-check times, process peak RSS, per-request latencies and raw module-power samples accompany each block.

The original plan specified 15W and 25W modes. Before any timing block, the 15W command returned a GPU-context reboot requirement; no reboot occurred and 25W was not attempted. The existing configuration was retained without reboot. Its nvpmodel label is MAXN\_SUPER/2, but actual TPC and FBP masks (both zero) differ from the installed definition (240 and 2). CPU bounds are 1.728\,GHz, GPU bounds 1.020\,GHz, and the EMC ceiling is 3.199\,GHz with override active. The costs apply to this recorded engineering-board configuration. No mode or clock changes occur during the 16-block measurement. Timestamped tegrastats and before/after snapshots preserve the observed state; the short blocks do not establish thermal steady state or absence of all throttling.

The continuous-inference benchmark cycles through the original 310 inputs with one outstanding request and includes complete CPU preprocessing and physical output materialisation. The continuous loop measures inference transformations without an application arrival schedule. Retained answers are used only for comparison; lookup service is not timed. Module energy includes CPU and GPU activity and is integrated over each 20\,s block without subtracting idle power. Raw integrations and latency counts were checked separately. The four groups repeat the comparison within one device session.

Median block mean module power was 16.58, 19.80, 12.80 and 17.51\,W for E, G, R and RG, respectively. The lower RG energy per request therefore does not imply lower average power. These are medians of integrated block means, not power limits. Local preservation is established by the saved post-block output banks; timed requests undergo finite/shape checks, without retaining every timed output for byte comparison.

%% file: si/edge_service.tex
\OpCertSInote{si:edge-service}{Heat-exchanger service energy and execution controls}

This note gives the complete-service measurements behind Figs.~\OpCertMainRef{fig:edge_sustained} and~\OpCertMainRef{fig:operating_cost}. The first comparison varies power and clock policy; the second compares manual reuse with standard freezing across request rates and service durations. They use different warmup schedules and remain separate experiments.

\subsubsection*{Power, clock policy and worker replacement}
The first comparison evaluates the same deterministic heat-exchanger checkpoint with a previously archived Jetson-local eager target. The complete raw CPU input and normalised and physical CPU outputs cross the worker boundary. E and RG share startup qualification, schema/finiteness checks and tracked-state guards. The comparison uses the fixed byte-identity criterion. Eight startup inputs are evaluated twice; 302 disjoint inputs are cycled during warmup and service. Each replicate starts a fresh worker.

The protocol records a 100\,ms response budget separately from the 100\,ms queue-age refusal threshold. Neither is inferred from an application deadline.

\begin{table}[!htbp]
\centering\small
\caption{\textbf{Heat-exchanger service across operating conditions.} Medians of three fresh-worker blocks per condition. Power covers arrivals; total energy covers preparation, warmup, arrivals, drain and closure, excluding bank loading and cooling. At 60\,Hz with automatic clocks, E and RG complete unequal work.}
\label{tab:edge-service-all}
\setlength{\tabcolsep}{4pt}
\begin{tabular}{@{}llrrrrr@{}}
\toprule
Policy & Arm & Rate (Hz) & Returned & p95 (ms) & Power (W) & Energy (J) \\
\midrule
15 W auto & E & 10 & 1200 & 21.97 & 5.549 & 841.9 \\
15 W auto & RG & 10 & 1200 & 8.07 & 5.202 & 799.4 \\
15 W auto & E & 60 & 5760 & 119.98 & 7.452 & 1071.2 \\
15 W auto & RG & 60 & 7200 & 7.15 & 5.952 & 889.6 \\
15 W fixed & E & 10 & 1200 & 11.76 & 5.960 & 944.7 \\
15 W fixed & RG & 10 & 1200 & 4.55 & 5.582 & 885.7 \\
15 W fixed & E & 60 & 7200 & 11.27 & 8.501 & 1250.0 \\
15 W fixed & RG & 60 & 7200 & 4.01 & 6.302 & 973.8 \\
25 W auto & E & 10 & 1200 & 21.80 & 5.524 & 838.6 \\
25 W auto & RG & 10 & 1200 & 8.32 & 5.176 & 796.6 \\
25 W auto & E & 60 & 5758 & 120.04 & 7.398 & 1065.0 \\
25 W auto & RG & 60 & 7200 & 7.16 & 5.927 & 887.3 \\
\bottomrule\end{tabular}
\end{table}

All 36 blocks completed the frozen protocol. Of 151,200 offered requests, 142,551 returned and 8,649 were refused by the queue-age policy. Every returned request was compared with the retained normalised and decoded target after delivery, with all recorded comparisons exact. The raw records retain the equality flags, executed source hashes, request timestamps, clock observations and module-power samples; they do not retain every full response array. The offline audit reconstructs request accounting and checks the executed comparison code and provenance; independent byte comparisons are limited to retained arrays. Temperature channels unavailable on the device remain marked unavailable. The earlier 16-block factorial retained complete post-block output banks.

The 15\,W fixed-clock, 60\,Hz blocks completed the same 7,200-request sequence in both arms. Paired mean-power reductions were 26.0968\%, 25.9088\% and 25.7282\%; preparation-to-closure reductions were 22.5289\%, 22.0427\% and 22.0126\%. We summarise the three paired ratios by their median. Neither cross-mode comparisons nor partially completed arrival sequences are pooled into this estimate. The 25\,W records form a separate session and provide an additional operating condition rather than a randomized isolated cap intervention.

The 12 traces contain 24 same-worker power-mode changes and 12 explicit replacement handoffs (Extended Data Fig.~\OpCertMainRef{fig:edge_recovery}). Replacement through qualification takes 3.779--3.850\,s and 22.18--22.65\,J after worker exit, and 4.565--4.630\,s and 26.81--27.30\,J after tracked mutation, across E and RG. These intervals are included in trace totals. Their 7,200 arrivals comprise 6,631 exact returned replies, 503 arrivals unavailable during recovery and 66 queue refusals. The controller detects the tracked mutation or worker exit during the control/request exchange and invokes recovery. The target is not regenerated. The older shear-flow controls in Note~\ref{si:reference-recovery} use separate fresh launches rather than measured handoffs.

The E/RG comparison holds the target, interface, guards and output ownership fixed. The earlier four-arm factorial in Supplementary Note~\ref{si:source-edge} separates reuse from graph replay at a different boundary; its component costs are not subtracted from this service. The compiler and Ray controls below test existing alternatives directly.

\subsubsection*{Compiler freezing of fixed-coordinate computation}
A separate mechanism probe uses the released deterministic heat-exchanger checkpoint and the same 310 input cases on RTX~5060~Ti. The wrapper registers the fixed coordinates as a buffer and passes only the two input branches to its forward method. PyTorch tracing followed by \texttt{torch.jit.freeze}, with \texttt{optimize\_numerics=False}, reduces the graph from twelve \texttt{aten::linear} operations to eight: the four trunk operations have been evaluated before inference. No OpCert-specific trunk rewrite is supplied to this compiler. Manual reuse and the frozen model both match the full eager output bytes on all 310 cases in each of three fresh processes.

The probe uses PyTorch \texttt{2.12.0.dev20260408+cu128}, one CPU thread, deterministic algorithms, TF32 disabled and \texttt{CUBLAS\_WORKSPACE\_CONFIG=:4096:8}. The input bank is normalised and resident on the GPU. After five warmups per path, 100 requests are timed with a device synchronisation after each request; the arm order is rotated between processes. The ranges of process-mean latency are 0.680--0.682\,ms for eager, 0.374--0.378\,ms for manual reuse and 0.354--0.358\,ms for compiler freezing. These timings exclude input transfer, decoding, worker transport and supervision; graph replay and power are not measured. The result shows that standard constant folding can eliminate the fixed trunk and retain exact agreement on this bank.

\subsubsection*{Request rate, service duration and preparation cost}
The extension fixes 15\,W automatic clocks and the original device-local target. Preflight banks from eighteen fresh workers (six paths, three workers each) reproduce all 310 normalised and physically decoded outputs byte for byte. The 90 heat-exchanger blocks comprise 36 six-path comparison blocks, 45 rate--duration blocks and nine ready-idle controls. All 153,189 offered requests return matching outputs; the nine ready-idle controls each include one request after the idle interval. Runtime agreement is recorded after delivery, whereas independent array comparison applies to the saved preflight banks.

\begin{table}[!htbp]\centering\small
\caption{\textbf{Matched heat-exchanger service with standard freezing.} Each session has three fresh workers per path at 40\,Hz over 30\,s; every block returns all 1,200 requests with matching outputs. Entries are medians. E, G, R and RG denote eager, graph, manual reuse and their combination; F and FG denote freezing without and with graph. Total energy includes preparation through closure.}
\label{tab:matched-freezing}
\begin{tabular}{@{}llrrr@{}}\toprule
Session & Path & p95 (ms) & Preparation (J) & Total (J)\\\midrule
1 & E & 21.28 & 21.26 & 228.78\\
1 & G & 16.86 & 22.00 & 225.55\\
1 & R & 11.46 & 20.24 & 185.35\\
1 & RG & 7.28 & 20.81 & 183.38\\
1 & F & 10.00 & 20.78 & 184.99\\
1 & FG & 6.83 & 21.61 & 183.74\\
\midrule
2 & E & 21.23 & 21.14 & 228.54\\
2 & G & 16.90 & 21.76 & 225.36\\
2 & R & 11.38 & 19.98 & 184.97\\
2 & RG & 7.20 & 20.63 & 183.03\\
2 & F & 10.09 & 20.57 & 184.64\\
2 & FG & 6.72 & 21.31 & 183.60\\
\bottomrule\end{tabular}
\end{table}

\begin{table}[!htbp]\centering\small
\caption{\textbf{Dependence on request rate and episode duration.} Three fresh-worker blocks per row at 15\,W with automatic clocks and no additional warmup. Each block completes all offered requests with matching outputs. Costs and response-time p95 are block medians. Frozen-artifact construction is excluded and reported separately.}
\label{tab:operating-lifetime}
\begin{tabular}{@{}lrrrrrr@{}}\toprule
Path & Rate (Hz) & Time (s) & Requests & p95 (ms) & Prep. (J) & Total (J)\\\midrule
G & 1 & 120 & 120 & 18.91 & 21.92 & 585.93\\
RG & 1 & 120 & 120 & 9.75 & 20.75 & 579.92\\
FG & 1 & 120 & 120 & 9.17 & 21.43 & 581.15\\
G & 10 & 10 & 100 & 17.70 & 21.82 & 77.95\\
RG & 10 & 10 & 100 & 7.88 & 20.65 & 73.06\\
FG & 10 & 10 & 100 & 7.50 & 21.38 & 73.68\\
G & 10 & 120 & 1,200 & 17.70 & 22.05 & 641.36\\
RG & 10 & 120 & 1,200 & 8.39 & 20.69 & 598.42\\
FG & 10 & 120 & 1,200 & 7.97 & 21.39 & 598.35\\
G & 10 & 600 & 6,000 & 18.08 & 21.87 & 3095.72\\
RG & 10 & 600 & 6,000 & 8.42 & 20.72 & 2886.48\\
FG & 10 & 600 & 6,000 & 7.96 & 21.30 & 2886.08\\
G & 40 & 120 & 4,800 & 16.85 & 21.86 & 822.05\\
RG & 40 & 120 & 4,800 & 7.40 & 20.76 & 657.26\\
FG & 40 & 120 & 4,800 & 6.81 & 21.46 & 656.85\\
\bottomrule\end{tabular}
\end{table}

\FloatBarrier
\paragraph{Ready-idle and construction costs.}
Median ready-idle power is 4.614, 4.608 and 4.611\,W for G, RG and FG, respectively ($n=3$, 60\,s each). These values measure module power, not power attributable to retained tensors. The frozen artifact contains eight Linear operations, compared with twelve before freezing. Five fresh-process builds consume 24.963, 19.806, 19.793, 19.616 and 19.716\,J in acquisition order. Their durations range from 3.425 to 4.451\,s. Each saved artifact is reloaded and reproduces all 310 archived normalised and physical output fields byte for byte. These measurements use the original builder and numerical settings, with the recorded 15\,W automatic-clock bounds. The interval starts before executor construction and ends after artifact saving; it includes NumPy/PyTorch imports, asset checks, input/model loading, CUDA initialisation, tracing/checking and freezing. Interpreter startup and the subsequent output audit lie outside the interval. A separate process samples module power at a nominal 20\,ms interval. Samples bracket every construction interval; the maximum gap is 20.554\,ms. Host filesystem caches are not cleared between processes. All five builds, including the slower first build, are retained.

Figure~\OpCertMainRef{fig:operating_cost} uses the five repeated builds; the preliminary acquisition is preserved separately in the source records. Charging their median, 19.793\,J, to each FG episode changes the median saving relative to G to $-2.50\%$ at 1\,Hz for 120\,s, $-20.01\%$ at 10\,Hz for 10\,s, $3.62\%$ at 10\,Hz for 120\,s, $6.13\%$ at 10\,Hz for 600\,s and $17.68\%$ at 40\,Hz for 120\,s. Negative values denote increased energy.

\paragraph{Measured saving available for artifact construction.}
For matched episode pair $g$ completing the same request sequence, define the energy saving before artifact construction is charged as
\begin{equation}
 \Delta E_g=E_{G,g}-E_{FG,g},\qquad
 B+E_{FG,g}<E_{G,g}\quad\Longleftrightarrow\quad B<\Delta E_g,
 \label{eq:si-build-margin}
\end{equation}
where $B$ is the separately measured artifact-construction energy. At 10\,Hz, the three paired savings span 4.096--4.277\,J over 10\,s, 42.502--43.045\,J over 120\,s and 208.706--213.614\,J over 600\,s. Every measured build (19.616--24.963\,J) exceeds every observed 10\,s saving and is below every 120 and 600\,s saving. The charged medians use the median build cost and three paired episodes. Figure whiskers span all combinations of the five observed costs and three episode pairs; these combinations are a cost sensitivity, not fifteen independent trials. Both service paths already include qualification and capture in preparation. The separate build charge applies to FG when the artifact is constructed; subsequent reuse of that artifact incurs no further build charge. It is not charged to manual reuse. This sensitivity uses the measured episode pairs and does not estimate a crossover time.

\paragraph{Structural controls and unavailable conditions.}
Darcy DeepONet uses its original $421^2$ checkpoint and normalised CPU interface; FNO uses its original $85^2$ checkpoint. Each block has a 120\,s arrival horizon, with three repeats per available path and rate (Table~\ref{tab:extension-structure}). The Darcy 1\,Hz G-to-RG paired reduction is 16.98\% (range 16.89--16.98\%). FNO E-to-G reductions are $-0.14\%$ at 1\,Hz (range $-0.19$ to $0.01\%$) and $0.49\%$ at 8\,Hz (range $0.33$--$0.53\%$). These total-service costs are separate from the earlier prepared-service cohort and are not pooled with its request-energy estimates.

The first Darcy 8\,Hz eager block returns 959 of 960 requests; one exceeds the 100\,ms queue-age limit. All returned outputs match. The equal-work requirement stops that queue. A continuation is stopped by the allocation-history guard during graph preparation after recorded allocation retries; it produces no additional timed block. The remaining eleven Darcy 8\,Hz blocks are unacquired. FNO completes in a separate continuation, with both arms acquired within that session. The extension therefore contains 115 acquired blocks, of which 114 complete equal work, and 162,068 matching replies from 162,069 arrivals. The refusal and preparation failure establish availability outcomes, not numerical drift.

\begin{table}[!htbp]\centering\small
\caption{\textbf{Darcy and FNO service costs.} All requests return matching outputs in each of three blocks per row at 15\,W with automatic clocks. Entries are medians; preparation-to-closure energy includes waiting. Absolute costs do not rank models with different grids. The incomplete Darcy 8\,Hz condition is described in the text.}
\label{tab:extension-structure}
\begin{tabular}{@{}llrrrr@{}}\toprule
Model & Path & Rate (Hz) & Requests & p95 (ms) & Total (J)\\\midrule
DeepONet & E & 1 & 120 & 154.24 & 704.81\\
DeepONet & G & 1 & 120 & 154.01 & 709.23\\
DeepONet & R & 1 & 120 & 31.44 & 588.79\\
DeepONet & RG & 1 & 120 & 33.01 & 588.82\\
FNO & E & 1 & 120 & 25.98 & 620.11\\
FNO & G & 1 & 120 & 24.71 & 620.72\\
FNO & E & 8 & 960 & 24.92 & 675.69\\
FNO & G & 8 & 960 & 23.16 & 672.51\\
\bottomrule\end{tabular}
\end{table}

\paragraph{Verification and measurement scope.}
Independent integration reproduces recorded block energies to within $2.1\times10^{-11}$\,J. The maximum sample gap in service blocks is 73.7\,ms and the highest recorded temperature is $50.8\,^{\circ}\mathrm{C}$. Sensor calibration uncertainty was not measured, so the approximately 1\% low-load differences remain observed instrument readings. CUDA memory metadata was sampled before witness qualification and records that snapshot, not a service peak.

\paragraph{Full-model TensorRT probe.}
The full-network FP32 engine with TF32 disabled, optimisation level 3 and a 1\,GiB workspace successfully builds and evaluates the heat-exchanger inputs. Independent comparison of all 310 archived first-pass outputs finds zero byte-identical cases in either the normalised or physical representation. The largest physical-channel relative $L_2$ difference from the local target is $1.2959\times10^{-6}$. This is a difference from the reference prediction, not an error against physical truth. The recorded second pass repeats, but its arrays are not archived for an independent repeatability audit. The candidate fails the byte criterion; application acceptance and service energy were not evaluated. Note~\ref{si:contractfrontier} reports the separate live-gated, fixed-trunk Darcy engine.

\paragraph{Reconstruction and exclusions.}
Figure reconstruction uses the audited request counts, power integrations and paired ratios. Unsuccessful preliminary attempts remain in the records and are excluded from completed-block counts. Note~\ref{si:legacy} reports the separate configuration study.

\subsubsection*{Comparison with Ray Serve}
Ray Serve~2.49.2 and OpCert share the fixed-trunk graph backend, complete CPU input/output interface, target, startup witnesses and live guards. The application adapter supplies the same numerical qualification to each replica constructor; Ray's health checks trigger automatic replacement against the unchanged target \citep{ray2025serve}. The local Ray instance has four logical CPUs; its single GPU replica receives one logical CPU allocation, one ongoing request and one queued request. Deployment handles bypass HTTP. Both services use one PyTorch thread, without CPU affinity. Ray uses a 1\,s health-check period, a 10\,s timeout and its default graceful-shutdown policy; OpCert uses a 0.5\,s stop timeout. Each driver saves and checks the 310-input output bank and warms for 20\,s. Inputs 8--309 are cycled with one outstanding request and a 100\,ms dispatch-age cutoff. Arrival-to-response, dispatch-to-response and queue age are recorded separately.

The separate heat-exchanger comparison offers 7,200 requests over 120\,s in the 15\,W automatic-clock configuration. Table~\ref{tab:established_service_baseline} retains every block; Supplementary Fig.~\ref{sifig:service_comparison}a--c shows the three primary pairs.

OpCert completes the offered sequence; Ray's remaining arrivals exceed the 100\,ms dispatch-age limit. Queueing accounts for most of Ray's longer arrival-to-response interval. The mean-power comparison therefore concerns unequal completed work. All 37,765 returned request digests in the six primary blocks agree with the retained target, and all 1,860 saved normalised--physical output pairs pass independent byte comparison. Both implementations recover after the tracked mutation. The recorded-cadence experiment below shortens Ray's health-check period from 1 to 0.1\,s and matches the 0.5\,s stop-time limit; the resulting interruption intervals belong to that separate configuration.

Primary results describe these configured services, rather than the best achievable performance of either framework. The prospectively repeated pair followed an evidence-copy operation spanning approximately six seconds and 109.5\,MiB of local writes during the Ray acquisition. Exact remote read boundaries were not logged. The replacement decision preceded inspection of response and power results, and no evidence copies occurred during the replacement measurements. Available temperature channels were monitored with a $60\,^{\circ}\mathrm{C}$ stopping threshold. Startup energy includes Ray instance construction; cooling, initial driver input-file loading and final framework shutdown are outside the startup, service and fault-to-successful-response intervals.

\begin{table}[t]
\centering
\small
\begin{tabular}{lrrrrr}
\hline
Service & Returned/7,200 & \shortstack{Response age\\(ms)} & \shortstack{Round trip\\(ms)} & \shortstack{Module power\\(W)} & \shortstack{Recovery\\(s)} \\
\hline
\multicolumn{6}{l}{Primary comparison: fresh-worker blocks} \\
OpCert (pair 0) & 7200 & 6.79 & 6.71 & 5.94 & 4.42 \\
Ray Serve (pair 0) & 5391 & 113.59 & 21.61 & 6.39 & 11.67 \\
OpCert (pair 2) & 7200 & 6.86 & 6.78 & 5.94 & 4.44 \\
Ray Serve (pair 2) & 5388 & 113.48 & 21.61 & 6.38 & 11.76 \\
Ray Serve (pair 3) & 5386 & 113.47 & 21.61 & 6.38 & 12.01 \\
OpCert (pair 3) & 7200 & 6.90 & 6.82 & 5.93 & 4.42 \\
\hline
\multicolumn{6}{l}{Original pair retained for sensitivity} \\
OpCert & 7200 & 6.87 & 6.79 & 5.93 & 4.44 \\
Ray Serve & 5359 & 113.54 & 21.63 & 6.40 & 12.59 \\
\hline
\multicolumn{6}{l}{Additional unmatched run retained} \\
OpCert & 7200 & 6.86 & 6.78 & 5.93 & 4.42 \\
\hline
\end{tabular}
\caption{\textbf{Matched application logic under two serving frameworks.} Response age and round trip are block medians of arrival-to-response and dispatch-to-response times. Mean module power includes controller activity, output auditing and waiting. Recovery is explicit in OpCert and automatic in Ray with one-second health checks. Primary pairs use OpCert--Ray, OpCert--Ray and Ray--OpCert order; excluded and unmatched blocks remain shown.}
\label{tab:established_service_baseline}
\end{table}

%% file: si/resource_history_detail.tex
\subsubsection*{Allocator-cap mechanism and interpretation}
The cap matrix uses 0.75, 0.83, 0.90 and 1.10$P$ on each device, where $P=A_{\rm peak}+W_{\max}$ is the empirical pressure coordinate: peak allocated memory plus the largest executed convolution workspace in one uncapped forward. RTX~5060~Ti adds 0.85$P$, 0.80$P$ post hoc, and the Rayleigh--B\'enard calibration in GiB. Seeded and trajectory inputs retain separate references. Ordered candidate-plan, workspace and allocation records identify the same ordinal convolution block across cap conditions.

At the profiled Rayleigh--B\'enard convolution block, the higher-cap plan used 37,914,640 bytes of workspace. At 0.20\,GiB, eleven candidate workspace allocations were refused before a zero-workspace plan executed. The trace distinguishes attempted workspace allocations from the plan that executes; the last descriptor of a failed forward is not counted as an executed plan. Numerical settings change the discrepancy magnitude but do not make the two cap regimes identical (Table~\ref{si:tab-controls}). Across eight test windows the cap changed 97.2--99.9\% of output elements; for window 120, relative $L_2$ was $2.526\times10^{-4}$, while both predictions had one-step VRMSE 0.0242 against truth. Thus execution discrepancy and prediction error have different scales.

The complete 127-case checkpoint/device/cap matrix is retained in the source data. Each mark is one fresh process; the 0.80$P$ condition was added post hoc. Outcomes include differing, identical and absent outputs, as well as explicitly unrun cases. A recorded allocator refusal accompanies every differing output and some identical outputs; it is not by itself a numerical-failure diagnosis.

Pool retention and cache policy were compared over 27 fresh-process histories; Supplementary Fig.~\ref{sifig:memoryhistory} shows the 18 shear-flow histories. The cuDNN v8 integration maintains a thread-local plan cache \citep{pytorch2026convv8}; releasing unused allocator blocks does not necessarily clear it. Releasing blocks while retaining cached plans could leave locally repeatable restored outputs different from their historical references. Retaining the blocks preserved the target; disabling plan caching restored it in the tested cases. The seeded pressure diagnostic uses exact numerical inequalities and float64-converted digests, separately from the original-dtype logical-byte audit. A separate GH200 study used eight windows and 24 policy trials: local repeats missed all historical disagreements; cache-disabled and fresh-process policies each recovered 24/24 targets. Its original forward/init costs remain in the archived table and are not pooled with service measurements.

\subsubsection*{Requalification against the persisted target}
The requalification control uses three fresh processes each for native eager, captured execution and independent fresh-process controls. The history uses allocator caps of 0.702, 0.29835 and 0.702\,GiB; its records identify Python 3.11.15, CUDA 12.9, cuDNN version code 92000 and driver 580.173.02. These software versions apply to this control. All 24 released outputs matched the pre-intervention reference. Eager processes refused at low memory and failed requalification after restoration; the two later unsupervised calls repeated without new allocator refusals, but differed in 524,014 of 524,288 elements. Captured execution refused a 20\,MiB request-path allocation, then requalified in the same process. Fresh-process controls reached READY 1.10--1.19\,s after harness start and returned exact outputs. They are independent launches, not timed handoffs from the refused workers. These are numerical-history controls on the shear-flow checkpoint; its unresolved physical-output contract is discussed in Note~\ref{si:physical}.

\subsubsection*{External occupancy bounds the resource-history interpretation}

A separate process allocated and touched CUDA memory before a fresh worker was created. Each cycle recorded two high-memory reference outputs, two outputs from a new worker under occupancy, two outputs from that same worker after occupancy release, and two outputs from another fresh worker. No per-process allocator fraction was imposed. We fixed nominal pre-worker free-memory targets of 1,024, 640 and 448\,MiB, with three independent cycles at each target and device. Actual CUDA free-memory snapshots and cuDNN execution traces were retained.

On RTX~5060~Ti, all six cycles at 1,024 and 640\,MiB completed with unchanged outputs throughout. At 448\,MiB, all three new workers failed during their first inference with an out-of-memory error; no pressured output was produced and the remaining calls were not run. All nine A2000 cycles completed with unchanged outputs. A2000 ran through WSL/WDDM, so its CUDA allocation accounting is not evidence that the same physical VRAM headroom or residency was imposed as on Linux. Supplemental 640\,MiB cycles recorded driver memory telemetry; they are instrumentation controls, not additional independent evidence selected for numerical drift.

Fresh-worker external occupancy did not reproduce the numerical change in these trials. The allocator studies establish a controlled execution-history mechanism, without estimating how often it occurs under ordinary co-tenancy.

%% file: si/si_table_numerical_controls.tex
\begin{tabular}{@{}>{\raggedright\arraybackslash}p{0.22\linewidth}rrr>{\raggedright\arraybackslash}p{0.30\linewidth}@{}}
\toprule
Control & Relative $L_2$ & \shortstack[r]{Differing elements\\(of 262,144)} & \shortstack[r]{Allocator\\refusals} & Note \\
\midrule
Default settings & $7.75 \times 10^{-4}$ & 261,675 & 16 & 0.20\,GiB cap versus unrestricted; TF32 on, benchmark off \\
cuDNN deterministic & $7.75 \times 10^{-4}$ & 261,675 & 16 & 0.20\,GiB cap versus unrestricted; same output digest as default \\
Deterministic algorithms and CUBLAS workspace configuration & $7.75 \times 10^{-4}$ & 261,675 & 16 & 0.20\,GiB cap versus unrestricted; same output digest as default \\
TF32 off on both sides & $2.87 \times 10^{-6}$ & 257,327 & 5 & 0.20\,GiB versus 0.35\,GiB cap; magnitude reduced, bits still differ \\
\midrule
External process holds 14.8\,GiB; no allocator cap & 0 & 0 & 0 & Uncapped; byte-identical to the unrestricted output \\
\bottomrule
\end{tabular}

%% file: si/temporal_scope.tex
\OpCertSInote{si:temporal-scope}{Field reconstruction, missed updates and recovery}

This note distinguishes reuse of input-independent computation from holding an earlier prediction (Supplementary Figs.~\ref{sifig:temporal_scope} and~\ref{sifig:service_comparison}d--f). A predictor is fitted once, then held fixed for a 121-observation service replay and offline update policies over the 2,642-observation test record. These controls assess update availability separately from the main energy comparisons.

\paragraph{Chronological reconstruction of measured fuel-cell fields.}
We constructed a deployment reference from the released PEMFC data and MIMONet implementation of Kobayashi et al.~\cite{kobayashi2026mimonet}. The source measurements are the University of Seville parallel-serpentine fuel-cell dataset \citep{toharias2024fuelcelldata} distributed with the MIMONet data and code release \citep{kobayashi2026mimonetdata}. We selected the first normal-flow dynamic-load experiment, recorded on 9 February 2022, by session metadata before fitting the predictor. Its segment-current and temperature files each contain 18,008 fields. Their acquisition timestamps agree at every frame to the recorded millisecond precision and increase monotonically. No temporal interpolation or alignment by nearest timestamp was required. The median interval is 1.000\,s; the 1st and 99th percentiles are 0.998 and 1.002\,s. The recorded intervals, including acquisition jitter, are retained.

We preserved the released spatial preprocessing: absolute values of the measured $18\times18$ segment-current map, denoted $I$ in A/segment, and first-order interpolation of the $9\times9$ temperature map, $T$ in $^\circ$C, to $18\times18$. Current values undergo no area normalisation. A seed-0 permutation selected 32 fixed grid locations for both channels. For temperature, the input before normalisation is $x_T=SU T_{\rm native}$, where $U$ interpolates the 81 native measurements to 324 grid locations and $S$ selects the 32 inputs. These inputs draw on 65 native temperature channels. The other 292 derived-grid locations are excluded from the input; they are not independently withheld native sensors. The reconstruction target is the complete pair of processed fields at the same acquisition time. This is spatial reconstruction from a time sequence of measurements, rather than temporal forecasting.

The original frame-randomised checkpoint was not used for chronological evaluation. We divided the raw sequence at 70\% and 85\% of its frame count and removed the first 60\,s of each validation and test interval. To bound reference-construction cost, training and validation used every tenth remaining frame; testing retained every frame. This rule was fixed before model fitting. The resulting training set contains 1,261 frames, from 15:41:48.754 to 19:11:22.726; validation contains 264 frames, from 19:12:28.727 to 19:56:18.727; and testing contains 2,642 frames, from 19:57:28.728 to 20:41:29.745, all on 9 February 2022 in the source clock. The source does not specify a time zone. Per-channel means and standard deviations were calculated over training frames and spatial locations only. An additive $10^{-6}$ was applied to each standard deviation, as in the released implementation. Validation and test frames contributed neither normalisation statistics nor training gradients.

The network retains the released nonlinear MIMONet architecture: width 256, GELU activations, three trunk layers, four layers in each branch, four layers in the combined decoder, and 872,450 parameters. Each branch receives the 32 normalised values of one channel together with their two spatial coordinates, giving 96 inputs. The trunk receives 128 Fourier features from 64 three-dimensional frequencies, generated with seed 7 and scale 3. The branch features are multiplied before concatenation with the trunk features and nonlinear decoding. We used seed 0, Adam with learning rate $10^{-3}$ and weight decay $10^{-5}$, batch size 64, and cosine learning-rate decay to $10^{-6}$ over 200 epochs. Training minimised normalised mean squared error over all 324 output locations. The checkpoint with minimum validation mean squared error over the 292 withheld grid locations was retained. All 200 epochs completed within the prespecified 480\,s limit; epoch 128 (zero-based index 127) was selected. The selected checkpoint is then fixed for the deployment comparison.

\paragraph{Accuracy control and aggregation.}
A multioutput ridge predictor used the same 32 selected grid locations and both channels, giving 64 normalised grid values as input and 648 normalised field values as output. The network's coordinate inputs are constant across frames. Ridge therefore receives the same varying measurement information. For training-centred input and output matrices $X$ and $Y$, we fitted
\begin{equation}
 \label{eq:si-ridge}
 W_\alpha=\arg\min_W\left\{\|XW-Y\|_F^2+\alpha\|W\|_F^2\right\}.
\end{equation}
The input and output means were calculated from training frames. We selected $\alpha$ from $\{10^{-4},10^{-2},1,100\}$ using the same validation criterion and locations as for MIMONet; $\alpha=10^{-2}$ was selected. No hyperparameter or checkpoint was selected using test error.

For channel $c$, frame $i$ and evaluation locations $\mathcal P$, let $y_{ipc}$ and $\hat y_{ipc}$ denote the physical target and prediction after reversing the training normalisation. We report the mean of the per-frame relative errors,
\begin{equation}
 \label{eq:si-relative-error}
 \overline{L}_{2,c}(\mathcal P)=\frac{1}{N}\sum_{i=1}^{N}
 \frac{\left[\sum_{p\in\mathcal P}(\hat y_{ipc}-y_{ipc})^2\right]^{1/2}}
 {\left[\sum_{p\in\mathcal P}y_{ipc}^2\right]^{1/2}+10^{-12}},
 \qquad N=2642,
\end{equation}
and the pooled physical root mean squared error,
\begin{equation}
 \label{eq:si-rmse}
 \mathrm{RMSE}_c(\mathcal P)=
 \left[\frac{1}{N|\mathcal P|}\sum_{i=1}^{N}\sum_{p\in\mathcal P}
 (\hat y_{ipc}-y_{ipc})^2\right]^{1/2}.
\end{equation}
Relative errors in Eq.~\ref{eq:si-relative-error} are dimensionless fractions and are multiplied by 100 in the tables. We evaluate both the complete 324-point field and the 292 withheld grid locations. Temperature RMSE is reported because relative error on the Celsius-valued field depends on its temperature offset. The chronological split tests later measurements from one experiment; it does not establish transfer to another cell or operating experiment.

\begin{table}[!htbp]
\centering\small
\caption{\textbf{Chronological fuel-cell reconstruction accuracy.} All entries use the same 2,642 test frames and physical targets. Relative $L_2$ is averaged over frames; RMSE pools frames and the indicated locations. Both predictors use the same 32 selected grid locations and training-only normalisation. Temperature targets inherit the released spatial interpolation from 81 measured locations.}
\label{tab:fuelcell-chrono-accuracy}
\setlength{\tabcolsep}{4pt}
\begin{tabular}{@{}llrrrr@{}}
\toprule
Predictor & Locations & $I$ rel. $L_2$ (\%) & $T$ rel. $L_2$ (\%) & $I$ RMSE (A/segment) & $T$ RMSE ($^\circ$C) \\
\midrule
MIMONet & All 324 & 1.3298 & 0.05294 & 0.001454 & 0.03744 \\
Ridge & All 324 & 0.09917 & 0.01564 & 0.0001090 & 0.01063 \\
MIMONet & Withheld 292 & 1.3212 & 0.05311 & 0.001445 & 0.03756 \\
Ridge & Withheld 292 & 0.1028 & 0.01647 & 0.0001131 & 0.01120 \\
\bottomrule
\end{tabular}
\end{table}

The two fixed predictors provide different fresh-field errors for the update-policy comparisons below. Ridge is more accurate in both channels.

We then compared the unchanged MIMONet checkpoint with an execution that retains its trunk features at the fixed spatial grid. The branch computations, multiplicative fusion, nonlinear decoder and physical-output conversion were unchanged. In three fresh workstation processes, the complete normalised and physical arrays were equal element by element for all 2,642 test inputs in each process. Complete physical output arrays were retained for both executions; normalised-output equality was recorded during execution. The three processes repeat the execution comparison for the same checkpoint and finite sequence. Prediction accuracy is evaluated separately against the processed measurement fields.

\paragraph{Observed-cadence serving and held-field assessment.}
Before the device trials we selected the first 120\,s of the chronological test sequence: 121 frames with their original timestamps. A device-local eager execution supplies an immutable reference bank and eight disjoint validation witnesses. Every driver receives the unchanged reference SHA-256. Both services use the same cached-trunk backend and complete CPU fields; branch computations, nonlinear decoding and numerical checks remain request dependent. All 121 outputs are checked before timing, so admission and evaluation concern this finite input population.

Three fresh-driver pairs replay the recorded elapsed times. A tracked mutation precedes the first request at or after 60\,s, observation 61 at 60.998\,s. OpCert replaces in a background thread so arrivals remain observable. Ray uses a 0.1\,s health-check period, 1\,s check timeout and 0.5\,s graceful-shutdown limit; OpCert's stop timeout is also 0.5\,s. One request is outstanding, with a 0.5\,s request timeout. Missed observations are not replayed, and cancellation is requested after a Ray timeout. Recovery ends at the first successful scheduled reply and includes the wait for that observation.

OpCert misses indices 61--64 and Ray misses 61--67 in each repeat. All 693 returned physical fields independently match the frozen arrays. Combined normalised--physical digests, twelve accepted admission records, distinct worker identities and common executed-source hashes also pass audit. Every returned prediction arrives before the next observation; response-time p95 ranges are 15.98--16.48\,ms for OpCert and 29.93--30.56\,ms for Ray. Available mode and temperature records remain within the protocol. 

The consumer assessment is performed after replay. It retains the latest returned field and evaluates it immediately before the next observation, including the final boundary at 120.998\,s. Neither service publishes held fields as new predictions. The acquisition cadence defines current delivery, not a plant safety deadline. All 121 intervals have a prior delivered field, so both implementations and all repeats use the same error population. The same missing-index sets recur in each replay, giving identical field-error scores (Table~\ref{tab:temporal-service}). The following analysis defines information age, separates the error terms in Eq.~\ref{eq:si_delivered_field}, and integrates energy over the common observation window.

\begin{table}[!htbp]
\centering\small
\caption{\textbf{Delivery and field error on the recorded fuel-cell segment.} Three repeats per service yield the same delivered sequence. Errors pool 121 observation intervals and 292 withheld grid locations. Segment-current magnitudes are in mA/segment and temperature in $^\circ$C. Time and energy entries are observed ranges over three replays. Energy includes replacement and waiting but excludes initial preparation and shutdown. The consumer holds the latest field between replies.}
\label{tab:temporal-service}
\begin{tabular}{@{}lrr@{}}
\toprule
Quantity & OpCert & Ray Serve \\
\midrule
Predictions returned / 121 & 117 & 114 \\
Observations without a current prediction & 4 & 7 \\
Maximum field age (s) & 5 & 8 \\
Fault-to-first-reply interval (s) & 4.006--4.008 & 7.005--7.010 \\
Module energy over 120.998\,s (J) & 612.308--613.437 & 632.421--633.427 \\
Segment-current RMSE from fresh prediction & 0.06112 & 0.08006 \\
Temperature RMSE from fresh prediction & 0.005120 & 0.007902 \\
\midrule
Segment-current RMSE against measured map & 1.481445 & 1.482078 \\
Temperature RMSE against interpolated map & 0.032113 & 0.032643 \\
\bottomrule
\end{tabular}

\smallskip
\parbox{.94\linewidth}{The corresponding fresh-model RMSEs are 1.481550\,mA/segment and 0.031686\,$^\circ$C. A held field can partially offset an existing prediction error: the OpCert current RMSE is marginally lower than the fresh-model RMSE. Thus freshness and physical accuracy do not have a monotonic relation on this sequence. Temperature targets inherit the original spatial interpolation of the measured $9\times9$ map.}
\end{table}

\subsubsection*{Refresh policies and observation-window energy}

The offline analysis uses the complete chronological test record described above: 2,642 observations spanning 2,641.017\,s. MIMONet and the more accurate ridge predictor remain fixed. The protocol identifies saved physical fields, measurements, timestamps and sensor indices by hash. The three host process banks agree elementwise. They are distinct from the Jetson service target: maximum differences on the first 121 fields are $2.98\times10^{-8}$\,A/segment and $7.63\times10^{-6}$\,$^\circ$C. Offline timing policies are evaluated separately from device service.

\paragraph{Prediction error and the available field.}
For one channel at common evaluation locations, let $u_i$ be the processed measurement field at observation time $t_i$, $r_i$ the selected predictor's fresh decoded field and $c_j$ the implementation output for observation $j$. In the device replay, immediately before $t_{i+1}$ a hold-last consumer has the latest delivered index $j=j(i)$; $u_i$ represents interval $i$, not a continuously observed field at its endpoint. The error decomposes as
\begin{equation}
 c_j-u_i=(r_i-u_i)+(c_j-r_j)+(r_j-r_i).
 \label{eq:si_delivered_field}
\end{equation}
The terms are fresh prediction error, implementation difference for the same input, and the change between the earlier and current reference predictions. Their squared norms contain cross terms, so reducing one term need not reduce measured-field error. Fixed-trunk reuse changes the work used to compute a fresh field; holding an output changes its observation index.

At this consumer boundary, information age is $t_{i+1}-t_j$ \citep{kaul2012realtimestatus}, so even a current field is about one acquisition interval old. The offline refresh analysis instead samples at $t_i$, where a fresh prediction has zero age. These age conventions are not pooled. The replay assesses delivered estimates without operating the fuel cell or a downstream controller.

\paragraph{Aligned prediction errors.}
Physical RMSE compares the field available under a policy with the corresponding processed measurement map over the 292 locations excluded from the inputs. A second RMSE compares it with the fresh prediction from the same model. Each channel uses its physical units. RMSE pools included timestamp--location pairs rather than averaging per-frame RMSE; relative $L_2$ is averaged over frames. Duration-weighted RMSE weights each frame's mean squared error by the next recorded interval, then takes the square root of the weighted average. It excludes the final frame when no next timestamp is available.

\paragraph{Refresh and age.}
The primary policy supplies a fresh prediction at indices $0,k,2k,\ldots$, for $k=1,2,4,7,10,20,30,60$, and retains it otherwise. Each policy is evaluated on all 2,642 observations. Acquisition spacing has median 1.000\,s and range 0.653--1.348\,s in this segment; $k$ is an observation count, not a strict wall-time period. All $k$ starting phases are additionally evaluated on the common index-60-onward population, using the available initial history. Separately, requested ages $a=0,1,2,4,7,10,20,30,60$\,s select the latest observation at or before $t_i-a$. Integer milliseconds avoid floating-point boundary errors. This comparison contains 2,581 observations beginning at index 61 (60.998\,s), after the required history exists. Actual ages can exceed $a$ by one acquisition interval; the populations are not pooled.

\begin{figure}[!htbp]
\centering
\includegraphics[width=\linewidth]{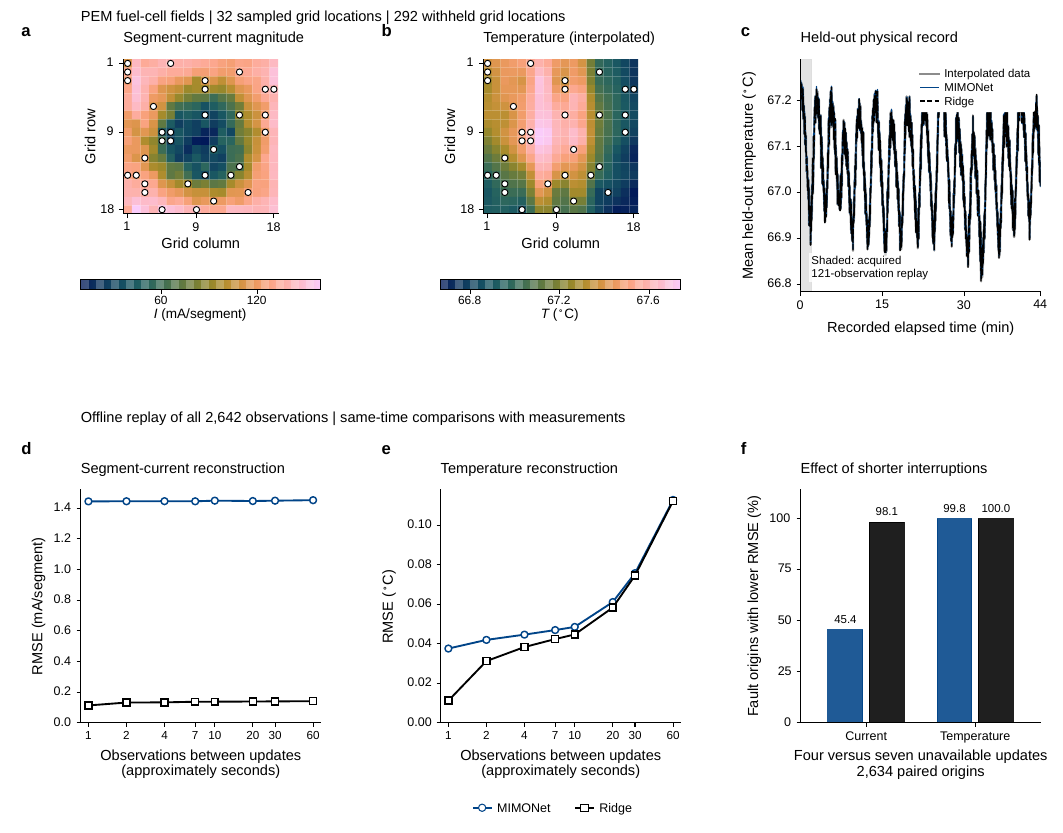}
\caption{\textbf{Holding an earlier prediction changes field error.}
\textbf{a,b}, Segment-current magnitude and interpolated temperature fields; markers locate 32 sampled grid locations.
\textbf{c}, Mean temperature at 292 withheld grid locations; shading marks the device-replay interval.
\textbf{d,e}, RMSE over 2,642 observations when predictions are refreshed every $k$ observations.
\textbf{f}, Percentage of 2,634 overlapping interruption origins with lower RMSE after four rather than seven missed updates. All policies are evaluated offline on the same record.}
\label{sifig:temporal_scope}
\end{figure}

At one update per 60 observations, MIMONet temperature RMSE increases from 0.03756 to 0.11279\,$^\circ$C, while segment-current RMSE changes from 0.001445 to 0.001453\,A/segment. Ridge temperature RMSE increases from 0.01120 to 0.11226\,$^\circ$C. Thus the large difference between fresh temperature predictors nearly disappears under this policy (Supplementary Fig.~\ref{sifig:temporal_scope}d,e). The omitted updates have no associated power measurement.

\paragraph{Interruption placement.}
For each eligible origin $j=1,\ldots,2634$, both policies retain prediction $j-1$ during interruption. One resumes after four unavailable updates and the other after seven. Both are scored over $j,\ldots,j+6$, with $t_{j+7}$ supplying the last interval boundary. Resuming fresh predictions for the final three observations cannot increase discrepancy from the fresh predictor, but error against measurements can increase or decrease. Shortening the interruption lowers MIMONet temperature error at 2,629 origins and segment-current error at 1,197; the corresponding ridge counts are 2,634 and 2,585 (Supplementary Fig.~\ref{sifig:temporal_scope}f). These percentages describe overlapping placements within this one record.

The median paired temperature-RMSE difference, seven minus four unavailable updates, is 0.00366\,$^\circ$C for MIMONet. Across nine consecutive 300\,s origin bins, including the shorter final bin, it ranges from 0.00334 to 0.00415\,$^\circ$C. Segment-current paired medians are negative in all nine bins, from $-4.49\times10^{-6}$ to $-1.68\times10^{-7}$\,A/segment. Measurement noise, model error and field variation are not independently identifiable from these maps; no safe interruption duration or diagnostic threshold is inferred.

\paragraph{Recorded-window energy.}
The device replay comprises the six blocks described above. Its original module-input samples are reintegrated from the first arrival through 120.998\,s, the next observation after the replay (Table~\ref{tab:temporal-service}). Monotonic timestamps align each power trace to this common window; bracketing samples permit linear boundary interpolation without extrapolation. Integration includes waiting, checks and the fault--replacement episode but excludes initial preparation, prechecking and final shutdown.

All six traces bracket both boundaries, and their maximum sampling gap is 23.18\,ms. Independently implemented clipped-segment integration reproduces the results. Alternative left- and right-held sample integration differs by at most 0.0144\,J. This check does not resolve sensor calibration uncertainty or loads outside the telemetry domain.

\begin{figure}[!htbp]
\centering
\includegraphics[width=\linewidth]{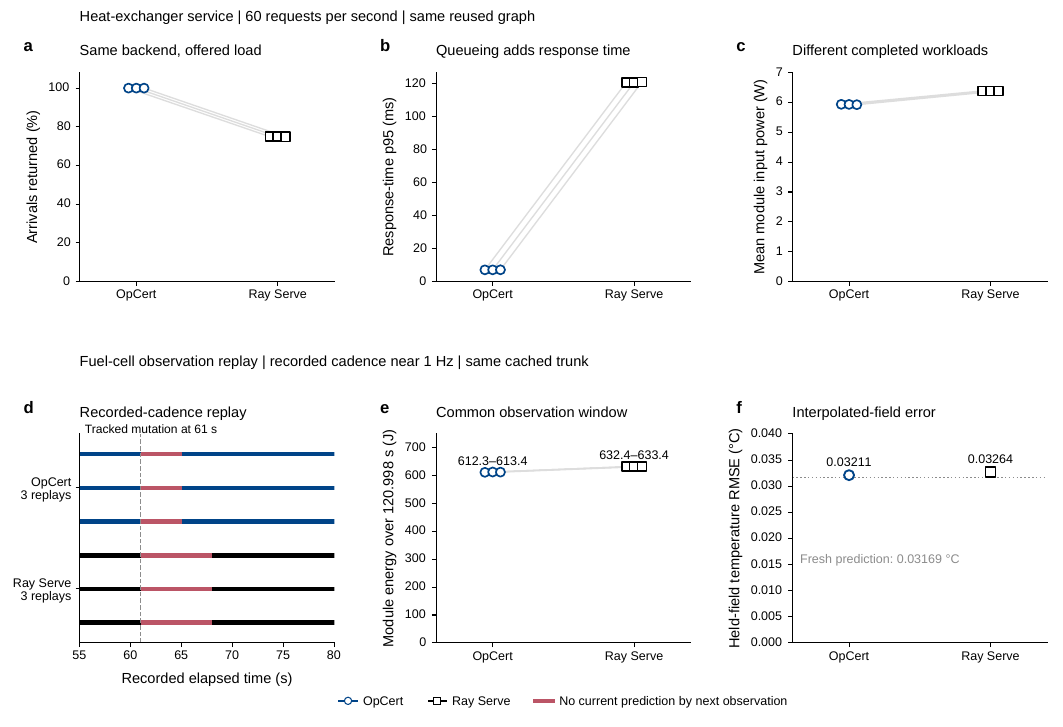}
\caption{\textbf{Service implementation affects cost and update availability.}
\textbf{a--c}, Heat-exchanger completion, response-time p95 and mean module power for three paired 120\,s blocks at 15\,W automatic clocks; each offers 7,200 requests.
\textbf{d--f}, Separate 121-observation fuel-cell replays: unavailable updates, module energy including replacement over 120.998\,s, and held-field temperature RMSE at 292 withheld grid locations. Lines pair three repeats; repeated delivery sequences give one RMSE per service. The dotted line marks fresh-prediction error. Within each comparison, services share the predictor and guards but complete unequal work.}
\label{sifig:service_comparison}
\end{figure}

Both services receive the same 121 observations but deliver different sequences. OpCert supplies three more updates while using 19.66--20.11\,J less sampled module energy in the paired blocks. The held temperature field is slightly more accurate against the measurements, while segment-current error remains approximately unchanged (Table~\ref{tab:temporal-service}; Supplementary Fig.~\ref{sifig:service_comparison}d--f). These are configured-service comparisons; transport and controller effects are not separately identified.

Policies and metrics were fixed before this reanalysis.

